\documentclass{article} 
\usepackage{collas2023_conference,times}
\usepackage{easyReview}
\usepackage{float}

\usepackage{amsmath,amsfonts,bm}

\def\eqref#1{equation~\ref{#1}}

\def\1{\bm{1}}
\newcommand{\train}{\mathcal{D}}

\def\va{{\bm{a}}}
\def\vb{{\bm{b}}}
\def\vc{{\bm{c}}}
\def\vd{{\bm{d}}}
\def\ve{{\bm{e}}}

\def\vh{{\bm{h}}}

\def\vk{{\bm{k}}}

\def\vq{{\bm{q}}}
\def\vr{{\bm{r}}}

\def\vu{{\bm{u}}}
\def\vv{{\bm{v}}}
\def\vw{{\bm{w}}}
\def\vx{{\bm{x}}}

\def\vz{{\bm{z}}}

\def\mA{{\bm{A}}}

\def\mE{{\bm{E}}}

\def\mI{{\bm{I}}}

\def\mO{{\bm{O}}}
\def\mP{{\bm{P}}}

\def\mW{{\bm{W}}}

\DeclareMathAlphabet{\mathsfit}{\encodingdefault}{\sfdefault}{m}{sl}
\SetMathAlphabet{\mathsfit}{bold}{\encodingdefault}{\sfdefault}{bx}{n}
\newcommand{\tens}[1]{\bm{\mathsfit{#1}}}

\def\tZ{{\tens{Z}}}

\def\gG{{\mathcal{G}}}

\def\gP{{\mathcal{P}}}

\def\sC{{\mathbb{C}}}

\def\sF{{\mathbb{F}}}

\def\sI{{\mathbb{I}}}

\def\sN{{\mathbb{N}}}

\def\sR{{\mathbb{R}}}

\usepackage{hyperref}
\hypersetup{
    colorlinks=true,
    linkcolor=red,
    filecolor=magenta,
    urlcolor=blue,
    citecolor=purple,
    pdftitle={Overleaf Example},
    pdfpagemode=FullScreen,
    }
\usepackage{booktabs}
\usepackage{multirow}
\usepackage{array}

\newcommand{\PreserveBackslash}[1]{\let\temp=\\#1\let\\=\temp}
\newcolumntype{C}[1]{>{\PreserveBackslash\centering}p{#1}}
\newcolumntype{R}[1]{>{\PreserveBackslash\raggedleft}p{#1}}
\newcolumntype{L}[1]{>{\PreserveBackslash\raggedright}p{#1}}
\usepackage{wrapfig,lipsum,booktabs}
\usepackage{enumitem}

\usepackage[margin=1in]{geometry}
\usepackage{subcaption}
\usepackage{amsmath,amsfonts,amssymb}
\newcounter{magicrownumbers}
\newcommand\rownumber{\stepcounter{magicrownumbers}\arabic{magicrownumbers}}

\title{Sample-Efficient Learning of Novel Visual Concepts}

\author{
Sarthak Bhagat$^*$, Simon Stepputtis$^*$, Joseph Campbell, Katia Sycara \\
The Robotics Institute, Carnegie Mellon University\\
\texttt{\{sarthakb, sstepput, jacampbe, sycara\}@andrew.cmu.edu} \\
}

\newcommand{\relate}[0]{\texttt{RelaTe}~}
\newcommand{\concept}[1]{\textit{#1}}

\collasfinalcopy 

\begin{document}

\def\thefootnote{*}\footnotetext{These authors contributed equally to this work}\def\thefootnote{\arabic{footnote}}

\maketitle

\begin{abstract}

Despite the advances made in visual object recognition, state-of-the-art deep learning models struggle to effectively recognize novel objects in a few-shot setting where only a limited number of examples are provided. Unlike humans who excel at such tasks, these models often fail to leverage known relationships between entities in order to draw conclusions about such objects. In this work, we show that incorporating a symbolic knowledge graph into a state-of-the-art recognition model enables a new approach for effective few-shot classification. In our proposed neuro-symbolic architecture and training methodology, the knowledge graph is augmented with additional relationships extracted from a small set of examples, improving its ability to recognize novel objects by considering the presence of interconnected entities. Unlike existing few-shot classifiers, we show that this enables our model to incorporate not only objects but also abstract concepts and affordances. The existence of the knowledge graph also makes this approach amenable to interpretability through analysis of the relationships contained within it. We empirically show that our approach outperforms current state-of-the-art few-shot multi-label classification methods on the COCO dataset and evaluate the addition of abstract concepts and affordances on the Visual Genome dataset.

\end{abstract}

\section{Introduction}
The ability to recognize objects from visual inputs \citep{Zhao2018ObjectDW} is crucial for the success of agents that interact in real or simulated environments \citep{9212350, NEURIPS2020_9909794d, robotics11060139}. Beyond applications in agent development, object recognition is also vital for image captioning~\citep{stefanini2022show}, scene understanding~\citep{XIE2020107205}, vision-language understanding~\citep{UPPAL2022149}, and many other domains. Recent contributions 
to foundational vision models \citep{dosovitskiy2021image, he2022masked} and a wider availability of computational resources has enabled many of these applications.
One benefit of such models is their ability to drastically reduce the amount of training data needed when utilizing them as priors to train new visual tasks, e.g. in the domain of object recognition.  
However, while very capable, these pre-trained models often fail to perform well in few-shot learning settings that require them to recognize novel objects from a small set of sample images \citep{wang2023worstcase}. Beyond object recognition, assigning abstract concepts and affordances is an even more challenging task as concepts such as \concept{wearable} are only indirectly related to a visual representation.
Inspired by how humans learn to utilize few-shot learning by connecting novel concepts to their prior domain knowledge and experience, neuro-symbolic architectures~\citep{Hassan2022NeuroSymbolicLP} can address some of these shortcomings by imbuing neural networks with symbolic knowledge graphs (KG)~\citep{AbuSalih2020DomainspecificKG}. 
Utilizing the interconnected domain knowledge of the graph, novel concepts can be added in a few-shot manner by augmenting the graph with the new nodes and thus, also limiting the need to re-train large parts of the neural architecture. Depending on the representation of novel nodes, such approaches are largely invariant to the topology of the graph, only requiring the final neural outputs to be expanded and trained while intermediate components may be able to only require little fine-tuning.
The availability of interconnected domain knowledge also allows easy integration of non-visual abstract concepts and affordances as relationships can be formed between such concepts and existing entities.

In the spirit of~\cite{Marino2016TheMY}, our approach utilizes a neural network approach in conjunction with an optimized KG constructed from the Visual Genome Multi-Label (VGML)~\citep{Marino2016TheMY} dataset. In this work, we improve and extend this setup to few-shot multi-label classification (FS-MLC) by proposing a pipeline that adds new information to existing domain knowledge via \relate: a multimodal relationship prediction transformer that, given a small set of images and a latent representation of the linguistic concept, automatically connects novel objects, abstract concepts, and affordances to existing domain knowledge.
In particular, \relate will evaluate the information propagated through the KG that is relevant to these images in the context of a latent concept representation from GloVe~\citep{pennington2014glove} and determine which nodes are applicable to be connected to the novel target concepts. Subsequently, we propose to extend the capacity of the final multi-label classifier by adding an output neuron associated with the new concept. The related weights of this extra neuron as well as the graph neural network are then trained and fine-tuned, respectively, to learn how to incorporate the new information. 
Thus, our approach utilizes a dynamically changing neuro-symbolic architecture that efficiently incorporates additional concepts in a sample-efficient manner.

Our proposed method provides improvements over existing work in two categories: 1) with the help of our proposed \relate module and training methodology, we show that it outperforms current state-of-the-art FS-MLC models on COCO and 2) our approach goes beyond object recognition by being able to also learn how abstract concepts and affordances can be incorporated in a few-shot manner, thus, continuously updating the neuro-symbolic architecture to accommodate the new concepts. 

The availability of a KG also makes this approach amenable to interpretability as the propagations throughout the graph can be reviewed after the classification is made~\citep{TIDDI2022103627}. 
The example in Figure~\ref{fig:overview} indicates that the \textit{motorcycle} concept caused the attribute \textit{two-wheeled}, \textit{red}, and the affordance \textit{transport}.
In summary, our main contributions are as follows: (a) We propose a sample-efficient few-shot methodology to recognize novel concepts from a small set of images by utilizing a knowledge graph that is amenable to interpretability, showing that it outperforms current state-of-the-art few-shot methods on the COCO dataset. (b) We introduce \relate -- a multimodal approach that predicts the existence of edges in the knowledge graph between novel concepts and existing nodes, allowing efficient integration of domain knowledge. (c) We also show the utility of having access to interconnected domain knowledge to effectively add abstract concepts, affordances, and scene summaries.
\section{Related Works}

Few-shot multi-label classification (FS-MLC) remains a challenging problem
despite some recent advances
~\citep{Alfassy2019LaSOLO, Chen2020KnowledgeGuidedMF, Yan_Zhang_Hou_Wang_Bouraoui_Jameel_Schockaert_2022}. On its own, few-shot classification is difficult due to various factors like catastrophic forgetting \citep{Goodfellow2013AnEI} and limited data sets; however, these problems are amplified in the multi-label case when novel target classes occur in conjunction with already existing concepts, making their identification and training even more challenging. One avenue of addressing this issue is the utilization of domain knowledge, which can reduce the complexity of this problem by reducing the reliance on labeled data~\citep{fsl_survey, Chen2021ZeroshotAF} and instead, drawing from the encoded knowledge. Such domain knowledge can be acquired in multiple ways, either by explicitly formulating and utilizing a data structure or by utilizing a foundational neural network that is ``large enough'' to encode the general knowledge. Examples of such large models are GPT~\citep{OpenAI2023GPT4TR}, particularly MiniGPT-4~\citep{zhu2023minigpt4}, CLIP~\citep{Radford2021LearningTV}, and Flamingo~\citep{Alayrac2022FlamingoAV}. However, in this work, we focus on imbuing neural networks with symbolic knowledge in the form of a Knowledge Graph as such data structure is amenable to human interpretation~\citep{guo2022explainable} and quick augmentation in order to address the FS-MLC problem. Nevertheless, we will compare our approach to publicly available large-language models (LLMs).






\paragraph{Few-Shot Multi-Label Classification}
Utilizing concepts has shown to be an efficient approach to learning interpretable policies~\citep{zabounidis2023concept}.  
One approach to learning the FS-MLC task is to define novel objects as the sum of their parts, allowing such approaches to learning how to recombine known, simpler concepts that represent the target class~\citep{Lake2011OneSL, Jia2013VisualCL}. However, the addition of novel fundamental concepts remains an active field of research. 
Several approaches have addressed the problem of adding new concepts from a small number of samples by utilizing additional modalities~\citep{Mao2015LearningLA,mei2022falcon}, structured primitives~\citep{Qian2019InferringSV}, generative modeling~\citep{Rostami2019GenerativeCC, Bhagat2020DisentanglingMF}, and meta-learning methods~\citep{cao21}. 
However, these approaches are usually limited to simulated (\cite{Mordatch2018ConceptLW, Qian2019InferringSV}) or less demanding datasets (\cite{Rostami2019GenerativeCC, Mao2019TheNC, cao21, mei2022falcon}) that do not reflect the richness and intricacy of real-world concepts that we encounter in our daily lives. 
One of the first papers addressing the problem of FS-MLC in great detail is~\citet{Alfassy2019LaSOLO}, which tackled the problem of limited data by representing sample images and their labels in a latent space and defining various set operations over these representations to synthesize additional samples through the combination of latent image features. 
Similarly, the work presented in~\citet{Yan_Zhang_Hou_Wang_Bouraoui_Jameel_Schockaert_2022} proposed a multimodal approach that utilized word embeddings to align verbal and visual representations in a latent feature space, allowing the creation of a mechanism that obtains visual prototypes for unseen labels by sampling an image from the latent space pinpointed by a language description of the novel entity.
In our work, we propose a framework for extracting abstract concepts from complex real-world images and demonstrate enhanced performance over current few-shot learners by utilizing the connection between linguistic and visual concept representations. 

\paragraph{Neuro-Symbolic Few-Shot Learning}
In addition to the techniques discussed above, incorporating domain-specific knowledge shows great potential as it can assist in recognizing and adding new concepts in a more sample-efficient manner, especially in scenarios with limited data. 
A commonly used approach to utilize symbolic KGs in deep learning is graph neural networks~\citep{Xie2019EmbeddingSK, Wan2019IterativeVR, Lamb2020GraphNN}, providing a multitude of benefits from interpretability~\citep{TIDDI2022103627} to the utilization of interconnected information.
The hierarchy and structure present in KGs have resulted in their use as priors for neuro-symbolic vision systems~\citep{Jiao2023GraphRL} for a number of applications ranging from transfer learning~\citep{alam_survey} to vision-language pre-training~\citep{alberts-etal-2021-visualsem}.
\citet{Chen2020KnowledgeGuidedMF} introduced a static knowledge-guided graph routing framework consisting of two graph propagation frameworks to transfer both visual and semantic features, enabling information transfer between correlated features to train a better classifier with limited samples. 
\citet{Marino2016TheMY} and~\citet{Fang2017ObjectDM} utilized this structured knowledge to identify the underlying concepts present in the image. With a comprehensive graph, the structured knowledge embedded in it can even be used to extract critical information about previously unseen classes in either a few-~\citep{Chen2019KnowledgeGT, Peng2019FewShotIR} or zero-shot~\citep{Kampffmeyer2018RethinkingKG, Wang2018ZeroShotRV, Huang2020MultilabelZC, 9607851, Wei2022SemanticEK, Lee2017MultilabelZL} manner.
However, a limitation of these approaches is the use of a static KG. The work presented in~\citet{wang} and~\citet{Kim2020VisualCR} partially addressed this problem by dynamically changing edge weights and re-computing latent node representations respectively, but the graph's structure and encoded knowledge fundamentally remain the same.
In this work, we propose a mechanism to update both aspects of the neuro-symbolic architecture by dynamically extending the KG with novel nodes, computing representations that are conditioned on the target images, and updating the neural components of our classification approach both structurally and in regards to its trained weights.
This also allows our approach to incorporate novel objects that go beyond the visual domain, including abstract concepts and affordances while alleviating the assumption, as in prior work~~\citep{Zhu2014ReasoningAO, 6942624, Chuang2017LearningTA, Ardn2019LearningGA, Gretkowski} that an exhaustive KG has to exist. 

\section{Few-Shot Object Recognition with Neuro-Symbolic Architectures}


In this work, we propose a twofold approach. Firstly, we employ a neuro-symbolic object recognition approach called Graph Search Neural Networks (GSNN), as originally introduced by~\citet{Marino2016TheMY}. To enhance the performance of this pipeline, we propose multiple modifications, namely adding image conditioning and incorporating node types (see Section~\ref{sec:nsor}). Secondly, we introduce a novel approach called \relate, which automatically extends the KG to integrate new concepts while, simultaneously, extending the neural components of the system to incorporate them (see Section~\ref{sec:concept_learning}). In the following sections, we provide detailed explanations of each component.

\subsection{Neuro-Symbolic Object Recognition}
\label{sec:nsor}
\begin{figure}[]
\centering
 \includegraphics[width=0.9\textwidth]{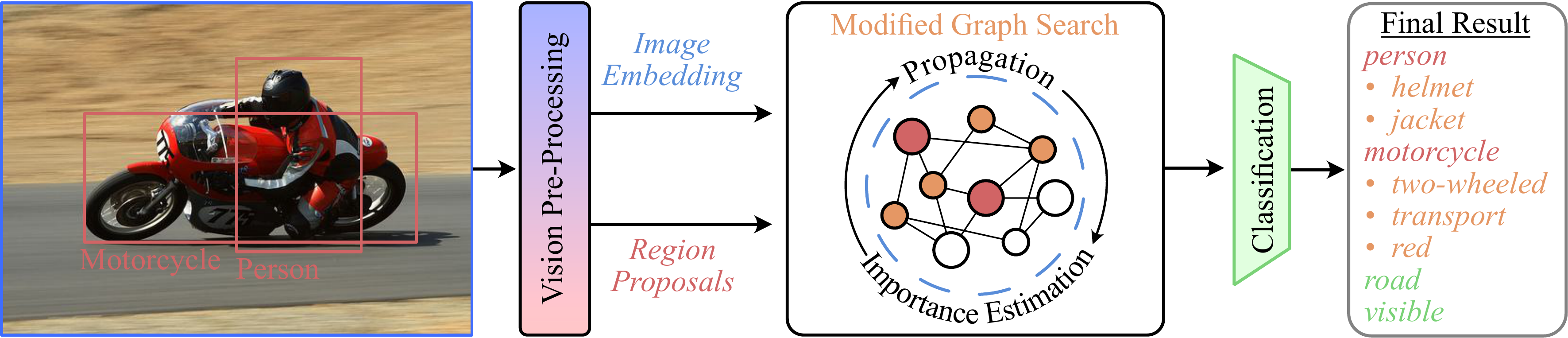} 
\caption{Example inference explaining our general inference pipeline, inspired by~\citet{Marino2016TheMY}. Given a novel image, we utilize ViT and Faster R-CNN to extract an image embedding (blue) and a set of initial object proposals (red). The initial proposals initialize our knowledge graph (red nodes) while the Modified Graph-Search identifies additional nodes (orange) conditioned on the overall image embedding (blue). Finally, our classifier evaluates the active nodes and produces a list of detected objects (green) in addition to the already detected nodes from the graph.} 
\label{fig:overview}
\end{figure}

At its core, our work considers the problem of detecting a set $\sC$ of concepts in a given image $\mI$, while affording the ability for a human Subject Matter Expert (SME) to extend the system's capability by detecting additional, novel concepts in a sample efficient manner from a small set of images. In this section, we first describe our inference pipeline, inspired by~\citet{Marino2016TheMY} before discussing novel concept addition in Section~\ref{sec:concept_learning}. Figure~\ref{fig:overview} describes the three main steps of the inference pipeline: First, we extract a set of candidate objects $\sF_\mI$ that initialize our KG $\gG$ and extract a global image embedding $\ve_\mI$ (see Section~\ref{sec:candidate_objects}); Second, we utilize the GSNN to propagate information through the graph while extending the prior work to also condition on the global image embedding $\ve_\mI$, alleviating the need for edge types, and utilizing semantic node types; Third, the final classifier evaluates all active nodes of graph $\gG$ (where $\gP_\mI$ is the sub-graph of $\gG$ containing the active nodes for a particular image $\mI$) in order to provide a holistic view of the input image and predict the final set of concepts $\sC_\mI$. 


\subsubsection{Extracting Candidate Objects}
\label{sec:candidate_objects}
In the first step, we employ a pre-trained object detection pipeline, namely Faster~R-CNN~\citep{Ren2015FasterRT} to extract the initial set $\sF_\mI$ of candidate objects from image $\mI$. 
Faster~R-CNN is pre-trained on the COCO~\citep{Lin2014MicrosoftCC} dataset to predict the $80$ concepts of COCO, but omit the $16$ classes designated for our FS-MLC experiments as defined in~\citet{Alfassy2019LaSOLO} for a total of $64$ trained concepts $\sC_{\text{COCO}}$. For this approach, we did not conduct any further fine-tuning on other datasets, nor did we change the outputs of Faster~R-CNN. The initial set of objects $\sF \subset \sC_{\text{COCO}}$ is then utilized to activate the initial set of nodes $\sN_\sF$ in graph $\gG$. 
In contrast to the prior work that uses VGG~\citep{simonyan2014very}, we use a pre-trained ViT~\citep{dosovitskiy2021image} model to calculate an overall image embedding $\ve_\mI \in \sR^v$ with feature size $v$ that is utilized to provide a global context for our modified graph-search approach as well as the final classifier. 
ViT is pre-trained on the ImageNet-$21$k~\citep{Deng2009ImageNetAL} dataset and then fine-tuned on the ImageNet-$10$k dataset without any further modifications.

\subsubsection{Modified Graph Search Neural Network}
\label{sec:GSNN}
\begin{figure}[]
    \centering
    \includegraphics[width=0.85\textwidth]{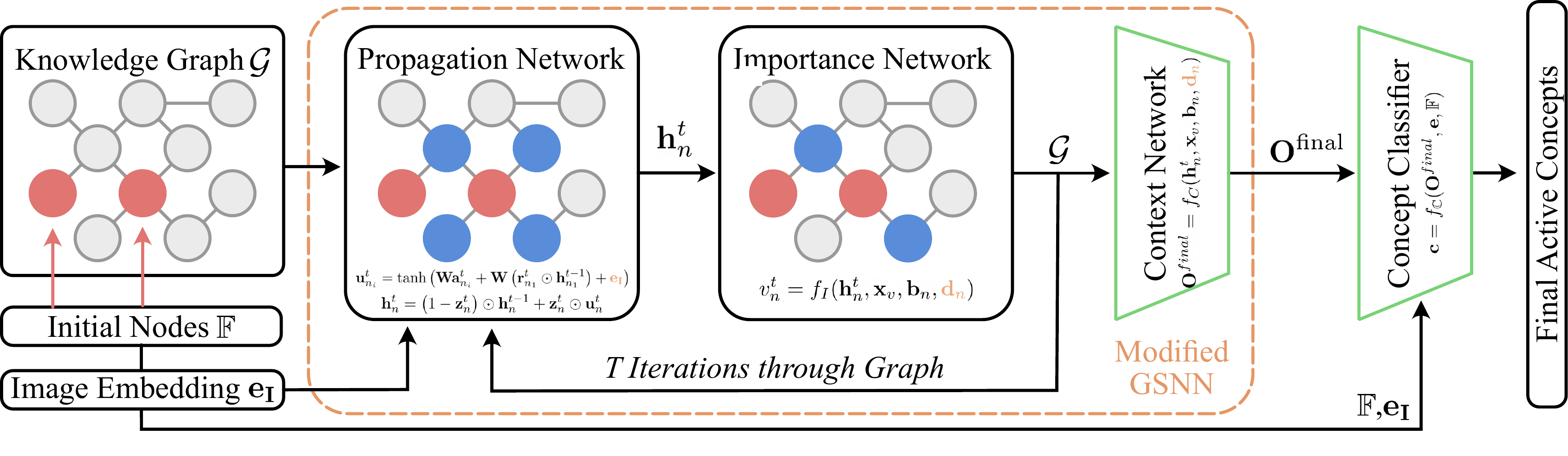} 
    \caption{Overview of the Modified Graph Search Neural Network (GSNN). In contrast to prior work, we condition the propagation network on the global image embedding $\ve_\mI$ and introduce a node type $d_n$ while dropping edge types. }
\label{fig:gsnn}
\end{figure}

In this section, we provide a detailed explanation of the different components of the GSNN inspired by \citet{Marino2016TheMY}, as well as the proposed modification of conditioning its components on the input image, removing edge types, and introducing node types.
Figure~\ref{fig:gsnn} shows the modified GSNN over graph $\gG$ which contains three core components: a)~the propagation network which computes an embedding for each node given its neighbors in the context of the current image $\mI$, b)~the importance network which decides which nodes are relevant and should be kept for potential future expansion given the current image $\mI$, and c)~the context network which generates final node embeddings. The context network is dependent on both the current image and the associations derived from the KG via multiple rounds of applying the propagation and importance network. The goal of the GSNN is to, in an iterative manner, propagate and prepare the information encoded in the KG that is relevant to image $\mI$ by alternating the propagation and importance network over $T$ steps. After $T$ rounds, the context network provides the representation needed for the final concept classifier. The selection of $T$ is a crucial hyper-parameter of our method and Section~\ref{sec:iterations} evaluates this choice in detail. Each of these components and the final classifier are described in the following sections along with our proposed modifications.


\textbf{Propagation Network:} 
Given the initial set of nodes $\sN_\sF$, the propagation network is designed to produce an output $\mO \in \sR^{N \times F}$, where $N$ is the number of nodes and $F$ the feature size for the latent node embedding, encoding the information of each node's neighborhood. 
Each row of $\mO$ represents a feature vector $\vh$ for the respective node which is initialized with all zeros outside of the first element, which contains the node ID, $x_v$.
We utilize the graph structure, encoded in an adjacency matrix $\mA \in \sR^{N \times N}$ to retrieve the hidden states $\vh$ of active nodes based on their neighborhood in the graph. 
In contrast to prior work, we also provide the propagation network with the global image encoding $\ve_\mI$ in order to ensure that information is propagated according to the image context (Ablations are provided in Section~\ref{sec:image_conditioning})

Initially, we calculate a vector $\va_n$ representing the neighbourhood of each active node at iteration $t$ given that $\mA_{n_i}$ is the relative adjacency matrix for node $n_i$: 
\begin{equation}
\label{eq:attention}
    \va_{n_i}^t = \mA_{n_i}^\mathsf{T}\left[\vh_1^{t-1}, \vh_2^{t-1}, \dots, \vh_N^{t-1}\right]^\mathsf{T} + b 
\end{equation}
Given this neighbourhood vector $\va_{n_i}^t$ for each node, we calculate $\vz_{n_i}^t$ and $\vr_{n_i}^t = \sigma\left(\mW \va_{n_i}^t + \mW \vh_{n_i}^{t-1}\right)$, where all $\mW$ are different and trainable weights of the neural network.
Subsequently, we calculate the update $\vu_{n_i}^t$ for each node's hidden state as follows, where each $\mW$ is a separately trainable weight matrix:
\begin{equation}
    \vu_{n_i}^t = \text{tanh}\left( \mW \va_{n_i}^{t} + \mW\left(\vr_{n_1}^t \odot \vh_{n_1}^{t-1}\right) + \ve_\mI \right)
\end{equation}
In contrast to the work presented in~\citet{Marino2016TheMY}, we calculate this update conditioned on the global image context $\ve_\mI$, allowing the modified GSNN to incorporate image-specific information. 
Section~\ref{sec:coco_baselines} evaluates this benefit in further detail. 
The final hidden state $\vh_{n_i}^t$ for each node in $\gP_\mI$ is subsequently calculated as a weighted sum of the previous hidden state $\vh_{n_i}^{t-1}$ and the previously computed update vector $\vu_{n_i}^t$:
\begin{equation}
    \vh_{n_i}^{t} = \big(1 - \vz_{n_i}^t\big) \odot \vh_{n_i}^{t-1} + \vz_{n_i}^t \odot \vu_{n_i}^{t}
\end{equation}
Together with the importance network detailed in the next section, the propagation throughout the graph is done over $T$ iterations, thus, allowing the utilization of the interconnected knowledge provided in the knowledge graph $\gG$. Learning to utilize the symbolic knowledge of the graph efficiently is of utmost importance for our few-shot learning goal described in Section~\ref{sec:concept_learning}.

\textbf{Importance Network.} 
The importance network alternates with the propagation network over $T$ cycles and decides whether or not an adjacent node to a currently active node should be made active.
This is an important step as purely expanding nodes at every step has the potential to become computationally impractical if $\gG$ is large.
The importance $v_n^t$ of each node at timestep $t$ is calculated as follows:
\begin{equation}
    v_n^t = f_I(\vh_n^t, \vx_v, \vb_n, \vd_n)    
\end{equation}
where, in contrast to the original GSNN, we propose the addition of $\vd_n$ which represents a one-hot vector describing the node type (``object'', ``affordance'', or ``attribute'') instead of using an edge type and $f_I(\dots)$ is a multi-layer perceptron (MLP). Nodes above a certain threshold $\gamma$ are maintained for the next propagation cycle. Additionally, we also learn a node bias term $\vb_n$ for each node in the knowledge graph that intuitively captures a global meaning of the respective node. Note that this bias is not depending on a particular image $\mI$.

\textbf{Context Network.} After $T$ iterations, the final node embeddings are created via the context network. 
Similar to the importance network, it is formulated as:
\begin{equation}
    \mO^{\text{final}} = f_C(\vh_n^t, \vx_v, \vb_n, \vd_n)    
\end{equation}
However, instead of predicting a scalar value indicating a node's importance, it generates the final state representation of the expanded nodes in $\gG$, where $f_C(\dots)$ is another MLP.

\subsubsection{Final Concept Classifier}
The third and final step is the classification of the active concepts $\sC_\mI \subset \sC$ in the input image $\mI$ where $\sC$ is a set of all possible concepts. These concepts are computed from the state representations $\mO^{\text{final}}$ of all the expanded nodes in the active graph $\gP$ along with the global image embedding $\ve_\mI$ and the originally detected classes $\sF_\mI$ from Faster~R-CNN. Utilizing a single fully connected layer, a probability distribution over all the concepts is predicted $\vc = f_{\sC}(\mO^{final}, \ve_\mI, \sF_\mI)$. In order to make the result amenable to interpretation by a human user, we also provide the graph of active nodes $\gP$, thus providing insights into why certain classifications may have been made.

\subsection{Novel Concept Learning in a Dynamic Neuro-Symbolic Architecture}
\label{sec:concept_learning}

In addition to improving the neuro-symbolic architecture of GSNN 
our remaining two main contributions are as follows: a)~a multi-modal Relation Prediction Transformer -- \relate, that aids a human SME when adding novel concepts to the symbolic knowledge graph (Section~\ref{sec:extension_relate}) and b)~introducing a framework to also dynamically updating neural parts of the inference pipeline described in Section~\ref{sec:nsor}.

\subsubsection{Extending the Knowledge Graph with Relation Prediction Transformer}
\label{sec:extension_relate}
\begin{figure}[]
    \centering
    \includegraphics[width=0.9\textwidth]{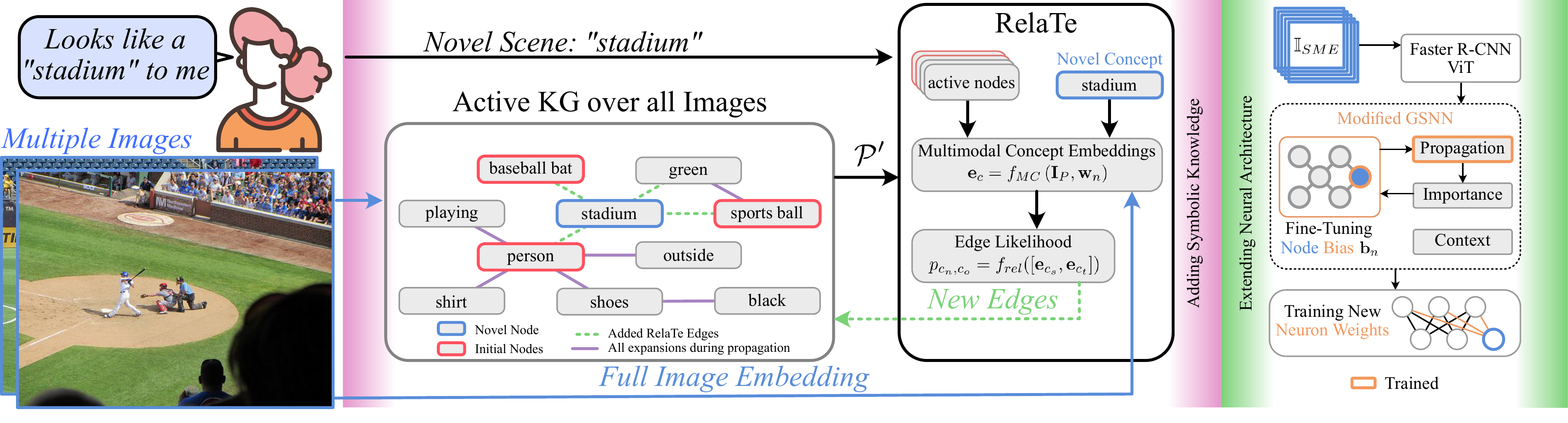}
    \caption{Given a novel concept of a \textit{stadium} by a human expert along with one or more images for it, \relate~estimates the optimal connectivity between the novel concept and existing nodes in the KG (i.e. \textit{person}, \textit{baseball bat}, \textit{sports ball}, \textit{green}). Subsequent inferences on similar images will yield the novel concept and allow for generalization through the domain knowledge encoded in the KG.} 
    \label{fig:relate}
\end{figure}

Figure~\ref{fig:relate} introduces our proposed approach -- \relate. Given a small set of SME-provided images $\sI_\text{SME}$ showing a novel concept as well as the partial graphs $\gP$ for each image, \relate predicts how this novel concept can be incorporated into the existing knowledge graph. Thereby, \relate provides an efficient and intuitive way of quickly adding new symbolic knowledge to the graph.

\textbf{Multi-modal Cross-Attention Framework.}
While we use the image processing pipeline of \citet{dosovitskiy2021image}, we introduce a multi-modal approach to relating linguistic concept representations to images that contain a novel concept. In particular, we utilize GloVe~\citep{pennington2014glove} word embeddings in order to retrieve a context-invariant representation $\vw_c = f_{GloVe}(c) \in \sR^{F_w}$ for any given concept $c \in \sC$. These representations are particularly useful when generalizing to novel concepts due to the potential similarity of new concept embeddings to a semantically similar word that may be known to the KG already. In order to combine the linguistic and visual representations, we utilize a cross-attention framework in which the image is represented as a sequence of patches $\mI_P$~\citep{dosovitskiy2021image, gan2022visionlanguage}.
We then combine both modalities as follows:
\begin{equation}
    \ve_{c_e} = f_{MC}\left(\mI_P, \vw_{n}\right) 
\end{equation}
where, $\vw_{c}$ is the word embedding of any given concept $c$, and $f_{MC}$ is explained in detail in Section~\ref{sec:cross_attn}.



\textbf{Post-Attention Fusion.} 
Given the embeddings $\ve_{c_e}$ for each node in each $\gP_\mI$ across all images in $\sI_{\text{SME}}$ and the novel concept $\ve_{c_n}$, we create pairs between the novel concept embedding $\ve_{c_n}$ and the existing nodes' embeddings $\ve_{c_e}$.
We calculate the likelihood of an edge being present between a source and target nodes by concatenating the embeddings of each node pair as follows:
\begin{equation}
    p_{c_s, c_t} = f_{rel}([\ve_{c_{s}}, \ve_{c_{t}}])
\end{equation}
Here, $f_{rel}(\dots)$ is an MLP predicting a scalar likelihood that an edge is present between the pair of nodes in the direction from source to target. $\ve_{c_{s}}$ and $\ve_{c_{t}}$ are populated by every combination of the novel concept node and existing graph nodes for novel objects; however, for abstract concepts and affordances, we only calculate incoming connections (see Section~\ref{sec:data} for details).
Finally, at most $k$ nodes that are above a specified threshold $p_{c_n, c_{o}} > \gamma$ are added to the KG. We empirically choose a suitable $k$ depending on the concept type while $\gamma$ is a global hyper-parameter.


\subsubsection{Updating the Neural Architecture}
\label{sec:few-shot-training}
Adding the novel concept to the KG alone does not directly yield improved classification performance as the new node does not have a trained node bias $\vb_n$ yet, nor does the propagation network know how to generate node embeddings $\vh_n$ for the novel concept. Additionally, the final classifier needs to be extended to enable the prediction of the novel class. In this section, we detail the process of training the node bias, fine-tuning the propagation module, and extending the classifier in further detail (also see Figure~\ref{fig:relate}, describing how we extend the neural architecture). 

\textbf{Fine-tuning the Node Bias and Graph Propagation.}
To fine-tune the propagation network and train the node bias $\vb_n$ of the novel concept, we utilize a small dataset $\train_{\text{SME}}$ generated from the images $\sI_{\text{SME}}$ given by the human SME that demonstrate the novel concept. $\train_{\text{SME}}$ is subsequently expanded by applying transformations to all images in $\sI_{\text{SME}}$. Further, we define a small curated dataset $\train_C$ with the intention of preventing catastrophic forgetting that contains $\sim2\%$ of the original VGML training data. The dataset $\train_C$ is selected through Maximally Diverse Expansion Sampling (MDES) which selects a representative set of inputs from the original VGML dataset that activates a diverse set of nodes in the graph $\gG$ (see Section~\ref{sec:mdes} for details).
Prior to training the propagation network and node bias, the novel node's bias $\vb_n$ is initialized as the average of its adjacent nodes' bias, and the corresponding novel node is forcefully activated in each training sample in $\train_{\text{SME}} \cup \train_{C} $ that contains an image from $\train_{\text{SME}}$. Forcing the activation of the novel node includes it in the downstream classification task and thus enables the fine-tuning of the propagation network and node bias.

\textbf{Extending the Classification Module} 
After training the propagation network and node bias for a limited number of epochs, the classification module with the novel neuron for concept $c_n$ is unfrozen and added to the fine-tuning process. Furthermore, we reduce the learning rate of the propagation module and freeze the node bias $\vb_n$ in this step of fine-tuning. As the classifier is depending on a valid node bias and the propagation network produces meaningful node embeddings for $c_n$, we delay the training of the classifier; however, we allow for continuous training of the propagation network to better capture the image-conditioned representation learning of the novel node.
When training the classifier, its training objective is reduced to a binary classification problem that predicts whether or not the novel concept is active in $\train_{\text{SME}} \cup \train_{C}$ while only calculating gradients for the added neuron so as to not alter the prediction capabilities of the existing concepts. This approach drastically reduces the number of parameters that need to be optimized, allowing the dataset to be comparatively small. 
\section{Evaluation}
We evaluate the effectiveness of our proposed approach for novel concept recognition in two separate settings. First, we compare it against current state-of-the-art FS-MLC baselines on  the COCO~\citep{Lin2014MicrosoftCC} dataset, and second, we perform a qualitative analysis of adding abstract concepts, affordances, and scene summaries to the underlying neuro-symbolic architecture. Particularly, we demonstrate that our method, when trained on Visual Genome, outperforms FS-MLC baselines on COCO and even improves performance further when trained on the COCO training data. Further, we conduct an extensive ablation study analyzing the impact of the various components of the neuro-symbolic architecture as well as our proposed \relate approach. Finally, we further investigate the implications of different node addition strategies, while additional experiments regarding the number of iteration step $T$, curated dataset $\train_C$, the addition of node types, and further qualitative analysis including deeper analysis in regards to failure cases and large language models are available in the Appendix. The source code can be found at: \texttt{https://github.com/sarthak268/sample-efficient-visual-concept-learning}. 

\subsection{Data and Metrics}
\label{sec:data}
As our method depends on the existence of an initial knowledge graph, we initialize the graph $\gG$ from the Visual Genome Multi-Label (VGML)~\citep{Marino2016TheMY} dataset. While based on Visual Genome~\citep{krishnavisualgenome}, VGML improves VG by drastically simplifying the graph, only using the $200$ most common objects and $100$ most common attributes, plus an additional $16$ nodes to completely cover all COCO classes.
We subsequently modify the graph for our FS-MLC task by removing $16$ nodes that are defined as test nodes in~\citet{Alfassy2019LaSOLO}, resulting in a total of $300$ nodes while also removing all images related to these $16$ FS-MLC target nodes from the training dataset of VGML. 
Furthermore, we impose the requirement that all nodes representing affordances and attributes must be leaf nodes in the graph to simplify graph structure further.
Additionally, we remove edge labels and introduce a one-hot vector indicating whether a node is an object, attribute, or affordance, and discovered that the edge types did not impact the performance of the approach. We evaluate these changes in comparison to the original knowledge graph from~\citet{Marino2016TheMY} in Section~\ref{sec:node_edge_ablation}. 

The coverage of all COCO classes is 
allowing us to compare novel object recognition performance between multiple COCO baselines and our approach. Particularly, the $16$ FS-MLC test classes include \textit{bicycle}, \textit{boat}, \textit{stop sign}, \textit{bird}, \textit{backpack}, \textit{frisbee}, \textit{snowboard}, \textit{surfboard}, \textit{cup}, \textit{fork}, \textit{spoon}, \textit{broccoli}, \textit{chair}, \textit{keyboard}, \textit{microwave}, and \textit{vase}.  In our additional evaluation of novel abstract concepts, affordance, and scenes, we utilize the full knowledge graph with all $316$ nodes.
We utilize our modified VGML dataset to train the GSNN, classification head of ViT, and final concept classifier in an end-to-end fashion. Further, \relate is trained on the entire Visual Genome dataset after removing the $16$ test classes from it. The training for \relate includes concepts that are not present in VGML.

\textbf{Evaluation Metric.} In order to compare the efficacy of our approach in the FC-MLC task, we utilize mean average precision (mAP), macro average precision (Macro AP), and the top-$K$ score. 
mAP is computed by taking the mean of the AP scores computed for each label, where AP is the area under the precision-recall curve plotted for each label.
Similarly, Macro AP is computed by averaging the AP scores for each label across all instances and then averaging the results across all classes.
To compute the top-$K$ score, we compute the percentage of $K$ most confident predictions of our model that are correctly predicted, i.e. precision of $K$ most confident predictions.


\subsection{Novel Concept Recognition}

In this section, we detail our comparison of utilizing \relate with our updated neuro-symbolic architecture to add novel concepts. Particularly, we compare against multiple state-of-the-art baselines on FS-MLC tasks over the COCO dataset while training our model on VGML and later fine-tuning on COCO. Further, we demonstrate the utility of adding scene summaries, e.g. \textit{kitchen} from kitchen appliances, abstract concepts, and affordances in a comprehensive ablation study on the VGML dataset. 

\subsubsection{COCO Novel Multi-Object Recognition}
\label{sec:coco_baselines}

We evaluate the efficacy of adding novel visual objects in a sample-efficient manner using our approach by comparing it against current state-of-the-art baselines in FC-MLC applications. As defined in~\citet{Alfassy2019LaSOLO}, we use a set $\sI_{SME}$ of five images per novel class and train all $16$ novel classes one by one. Table~\ref{table:coco_quant} shows the results comparing our method to three state-of-the-art baselines as well as four additional ablations. For each method, ``Source'' indicates the training dataset for the respective model while $\train_{SME}$ and $\train_{C}$ indicate which dataset was used for the few-shot learning. If both datasets are used, they are randomly interleaved. Lines 1 to 3 in Table~\ref{table:coco_quant} demonstrate the performance of our three baselines.
Using a naive approach to adding novel concepts to~\citet{Marino2016TheMY}, line 4 trains only the final classifier (by adding a novel neuron) on the same training dataset as used in lines 4 to 7 without adding novel information to the knowledge graph altogether. In addition to training the classifier, adding the novel node to the knowledge graph, but not training the propagation network and node bias is shown in line 5, indicating that our modified GSNN is mostly invariant of the knowledge graph despite the lack of fine-tuning, underlining the strength of having domain knowledge (compare line 4 and 5), yielding a $17\%$ performance increase. However, further improvements can be done when fine-tuning the propagation network and node bias.
Compared to~\citet{Yan_Zhang_Hou_Wang_Bouraoui_Jameel_Schockaert_2022} in line 3, which achieves $68.12\%$ (the best baseline), utilizing the neuro-symbolic architecture and our proposed \relate architecture in line 6, we achieve a Macro AP score of $70.26$, despite training on VGML, which is statistically significant with a standard deviation of $\sigma = 0.45$ at $p$-value $1.252e^{-3}$ trained over four seeds. 
Further, we also fine-tuned our method from line 6 on the training set of COCO and report the results of $70.30\%$ with $\sigma = 0.19$, with a $p$-value of $9.1e^{-5}$ over four seeds in line 7 of Table~\ref{table:coco_quant}. In each case, we parameterize \relate with an unlimited $k$ value to add as many relations as possible.
Given that our results in line 6 are resulting from a model trained on an entirely different dataset, i.e. VGML, yet performs very similarly to being trained on COCO allows the conclusion that our approach has the ability to transfer knowledge between datasets through the utilization of a knowledge graph.


\subsubsection{Recognizing Affordances, Attributes, and Scenes}
\label{sec:nonvisual}

\setlength{\tabcolsep}{3pt}
\setcounter{magicrownumbers}{0}
\begin{wraptable}{r}{6.6cm}
    \centering
    \scriptsize
    \begin{tabular}{cc|ccc|c}  
        \toprule
         & Method & Source & $\train_{SME}$ & $\train_{C}$ & Macro AP \\ 
        \midrule
         \rownumber & \cite{Alfassy2019LaSOLO} & COCO & \checkmark & --- & 58.10 \\
         \rownumber & \cite{Chen2020KnowledgeGuidedMF} & COCO & \checkmark & --- & 63.50 \\ 
         \rownumber & \cite{Yan_Zhang_Hou_Wang_Bouraoui_Jameel_Schockaert_2022} & COCO & \checkmark & --- & 68.12 \\
         \rownumber & Fine-tuning (classifier) & VGML & \checkmark & \checkmark & 52.22\\ 
         \rownumber & Fine-tuning + \relate & VGML & \checkmark & \checkmark & 69.26\\ %
         \rownumber & Ours & VGML & \checkmark & \checkmark & 70.26\\ 
         \rownumber & Ours & COCO & \checkmark & \checkmark & \textbf{70.30} \\
         \bottomrule
    \end{tabular}
    \caption{Experimental results on COCO dataset for five-shot multi-label classification of previously unseen concepts.}
    \label{table:coco_quant}
\end{wraptable}
Unlike other approaches to few-shot novel concept detection that rely on novel objects being visible in the input image, our approach can go beyond such limitations through the utilization of interconnected information in the knowledge graph. In addition to adding visual concepts as shown in Section~\ref{sec:coco_baselines}, we demonstrate how non-visible concepts like abstract concepts, attributes, and scene summaries can be added. While the borders between what is visual and what is not are sometimes blurry, particularly in the case of scenes, utilizing the knowledge graph highlight the ability to draw conclusions from a set of partial observations. E.g., given that \textit{refrigerator}, \textit{oven}, and \textit{microwave} were detected, we can conclude that the input image likely shows the \textit{kitchen} concept, which can subsequently be added to the knowledge graph as a novel concept. In the following two sections, we discuss the addition of abstract concepts and scene summaries. 


\paragraph{Adding Non-Visual Concepts.}

\begin{wraptable}{r}{7.5cm}
    \centering
    \scriptsize
    \begin{tabular}{cccccccc}  
        \toprule
        &  & \multicolumn{6}{c}{\textbf{Scene Concept}} \\ 
        & \textbf{Model} & \textit{stadium} & \textit{kitchen} & \textit{zoo} & \textit{school} & \textit{bedroom} & Avg.\\
        \midrule
        \rownumber & CLIP (0-shot) & 16 & 100 & 100 & 56 & 72 & 68.8 \\
        \rownumber & Flamingo (0-shot) & 20 & 4 & 40 & 24 & 28 & 23.2 \\
        \rownumber & Mini-GPT (0-shot) & 24 & 96 & 64 & 24 & 96 & 60.8 \\
        \rownumber & Flamingo (5-shot) & 68 & 36 & 40 & 72 & 80 & 59.2\\
        \rownumber & Ours (5-shot) & 90 & 84 & 84 & 72 & 92 & \textbf{84.4} \\
        \bottomrule
    \end{tabular}
    \caption{Novel scene prediction in comparison to free-form text generation models.}
    \label{table:scene}
\end{wraptable}
We conduct further experiments to assess the ability of \relate to incorporate non-visual concepts into the knowledge graph by relating it to relevant existing domain knowledge. In contrast to novel object recognition, we parameterize \relate with a threshold $k = 3$ in order to enforce a sparser connection of affordances and attributes to existing nodes. 
\begin{figure}[]
    \begin{subfigure}{0.49\linewidth} \centering
        \includegraphics[width=1\linewidth]{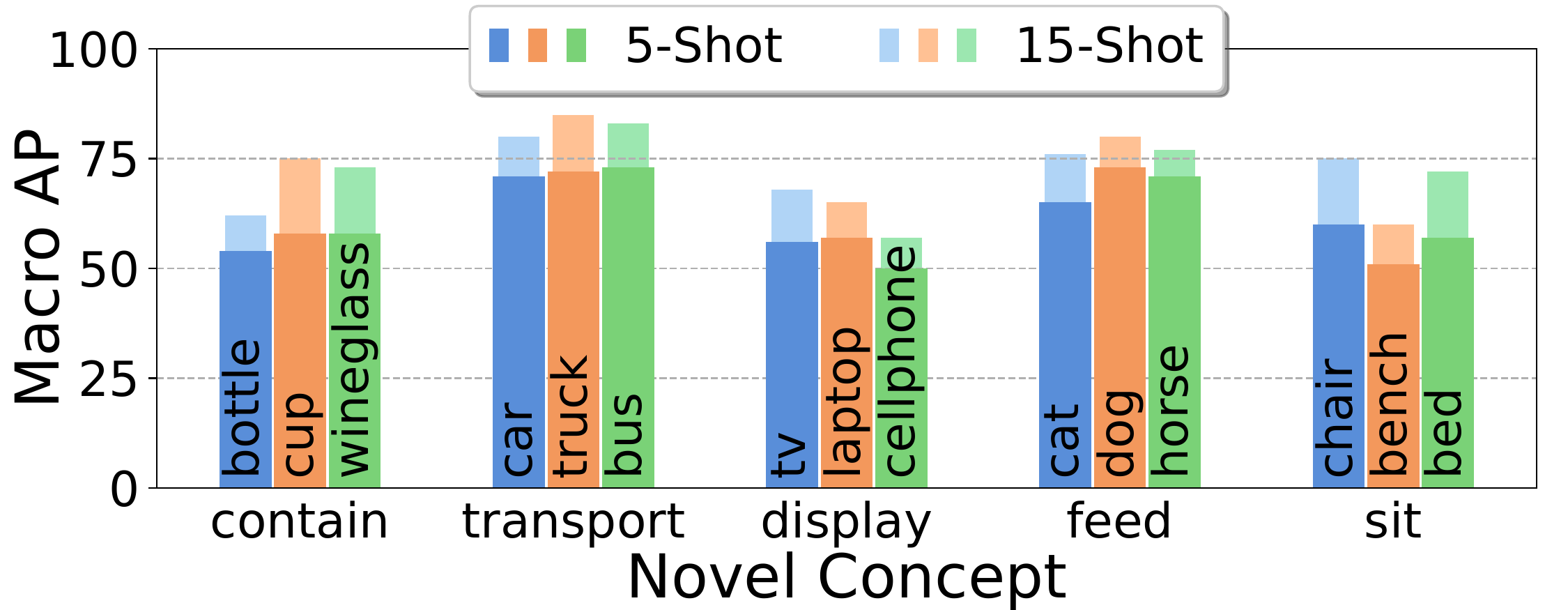}
        \caption{Novel affordances learning given a set of five images.}
        \label{fig:affordances}
    \end{subfigure}%
    \hspace{0.02\linewidth}%
    \begin{subfigure}{0.49\linewidth} \centering
        \includegraphics[width=1\linewidth]{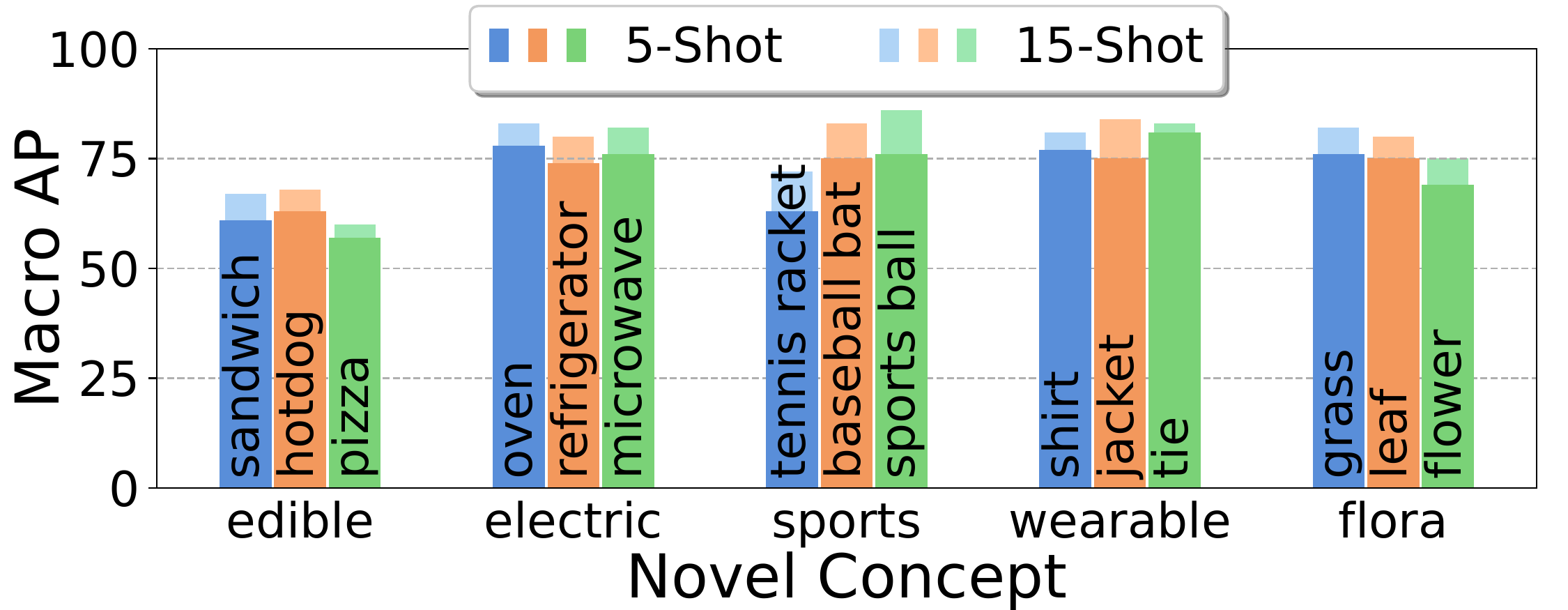}
        \caption{Novel attribute learning, given a set of five images.}
        \label{fig:attributes}
    \end{subfigure}
    \caption{Analysis of the performance when adding novel affordances and attributes to the knowledge graph. We evaluate the performance on five-shot (dark colors) and fifteen-shot (bright colors) learning.} 
    \label{fig:nonvisual}
\end{figure}
Figure~\ref{fig:nonvisual} shows our experiments on adding novel affordances (Figure~\ref{fig:affordances}) and attributes (Figure~\ref{fig:attributes}). In each case, we selected a set $\sI_{SME}$ with five and fifteen sample images containing three separate concepts that should be assigned to the novel concept. Subsequently, we evaluate the performance of the resulting model on $50$ test images from within the same concept classes as well as $50$ test images that do not show any of the trained targets. 
Our results show that added non-visual concepts have an average Macro AP of $66.7\%$ given five sample images and $75.1\%$ given fifteen images. Recall that for novel object detection as shown in Table~\ref{table:coco_quant}, the Macro AP score is $70.26\%$.
We hypothesize that the slightly lower performance on abstract concepts roots from the difficulty of not having a clear visual representation for such concepts. However, when increasing the training samples to fifteen, we outperform the object detection reported in Table~\ref{table:coco_quant}, which, in the context of deep learning, is still a relatively small sample size.

\paragraph{Novel Scene Recognition:}


\setcounter{magicrownumbers}{0}
In addition to adding novel objects and abstract concepts, \relate can also assist in the addition of compound concepts.
For example, the existence of a \textit{oven}, \textit{microwave} and \textit{refrigerator} implies a scene that can be defined as a \textit{kitchen} that is the sum of the underlying parts. 
Encoding such knowledge poses a slightly different problem as compound concepts require reasoning over multiple adjacent concepts. This ability can be imbued by our KG, but can also be found in large foundational neural networks, particularly LLMs.
Table~\ref{table:scene} demonstrates the ability of three LLM baselines, MiniGPT-4~\citep{zhu2023minigpt4}, CLIP~\citep{Radford2021LearningTV}, and Flamingo~\citep{Alayrac2022FlamingoAV}, to draw higher-level conclusions about the general scene shown in an image to our method, attempting the same task.
We evaluate each scene on $25$ test images of previously unseen samples and report the existence of the compound concept within the estimated concepts. For the LLMs, particularly the free-form response models, we queried the models if the image shows any of the five target scenes. 
With an average accuracy of $84.4\%$, this experiment underlines the utility of having interconnected knowledge that augments our few-shot detection pipeline, allowing the GSNN to successfully draw high-level conclusions from a set of basic concepts. 
While LLMs demonstrate partial success in identifying the high-level scenes, explicitly modeling the symbolic knowledge provides significant improvements despite the LLM's large general knowledge encoded within their trained model architecture.
In additional experiments for the \textit{kitchen} example, we showed that the likelihood of classifying the \textit{kitchen} scene from a \textit{refrigerator} or \textit{microwave} alone is $24\%$ and $28\%$ respectively while the likelihood to identify it from an image containing both base concepts is $88\%$, showing that our model accurately learned that a \textit{kitchen} is the sum of its parts. 


\subsubsection{Ablations}
\setcounter{magicrownumbers}{0}
\setlength{\tabcolsep}{3pt}
\begin{table*}[h]
    \scriptsize
    \centering
     
    \begin{tabular}{cccc|cc|cc|ccc|ccc} 
        \toprule
        & \multicolumn{3}{c|}{\textbf{Components}} & \multicolumn{2}{c|}{\textbf{Fine-tuned}} & \multicolumn{2}{c|}{\textbf{KG Configuration}} & \multicolumn{3}{c|}{\textbf{All Classes}} & \multicolumn{3}{c}{\textbf{Novel Classes}} \\ 
        & ViT & FRCNN & KG & ${\tt GSNN}$ & CLF & ${\tt RelaTe}$ & MDES & \textbf{T-1} & \textbf{T-5} & \textbf{mAP} & \textbf{T-1} & \textbf{T-5} & \textbf{mAP}\\ 
        \midrule
        \rownumber  & \checkmark & - & - & - & - & - & - & 84.2 & 63.4 & 31.8 & 85.7 & 65.5 & 34.6\\ 
        \rownumber  & \checkmark & \checkmark & - & - & \checkmark & - & - & 84.7 & 64.2 & 33.0 & 86.1 & 67.2 & 37.3\\
        \rownumber  & \checkmark & \checkmark & \checkmark & - & \checkmark & - & - & 87.4 & 68.8 & 36.5 & 91.8 & 72.8 & 68.0\\
        \rownumber & \checkmark & \checkmark & \checkmark & - & \checkmark & \checkmark & - & 89.8 & 69.8 & 38.5 & 91.6 & 72.8 & 68.2\\ 
        \rownumber & \checkmark & \checkmark & \checkmark & \checkmark & \checkmark & \checkmark & - & \textbf{90.4} & 72.4 & 41.7 & 92.2 & 73.6 & 69.3 \\ 
        \rownumber & \checkmark & \checkmark & \checkmark & - & \checkmark & \checkmark & \checkmark & 90.0 & 70.1 & 39.5 & 91.8 & 73.0 & 68.8\\ 
        \rownumber & \checkmark & \checkmark & \checkmark & \checkmark & \checkmark & \checkmark & \checkmark & 90.3 & \textbf{72.9} & \textbf{42.0} & \textbf{92.4} & \textbf{73.9} & \textbf{69.5} \\ 
        \bottomrule
    \end{tabular}
    \caption{Experimental results on Visual Genome dataset, ablating the components of our method}
    \label{table:vg_quantitative}
\end{table*}

In this section, we ablate the different components of our FS-MLC pipeline on the VGML dataset, recognizing the novel objects defined in~\citet{Alfassy2019LaSOLO}. Table~\ref{table:vg_quantitative} summarizes these results where lines 1 and 2 show the performance on the test set across all classes and our 16 novel few-shot classes given their Top-1 (T-1) and Top-5 (T-5) performance when using pure neural end-to-end architectures. In each case, novel classes are trained with five demonstration images and evaluated on the test set of VGML. Line 3 adds a KG with the GSNN approach proposed in~\citet{Marino2016TheMY} and fine-tunes the final classifier (CLF-column) on the novel classes with an $\sim3\%$ improvement in Top-K score. From this, we conclude that novel classes may also need to be added to the knowledge graph. Line 4 uses our proposed \relate approach to add the novel classes to the graph; however, does not tune the GSNN with respect to the propagation network and node bias (GSNN-column). 
Adding nodes to the graph yields another $3-5\%$ improvement over line 3. Line 5 fine-tunes the propagation network and node bias with our methodology described in Section~\ref{sec:few-shot-training}, improving performance by another $\sim2\%$. Finally, lines 6 and 7 show the impact of our curated fine-tuning dataset $\train_C$ as compared to an equally sized random dataset over the original VGML dataset. This demonstrates the importance of MDES to prevent catastrophic forgetting.
In summary, Table~\ref{table:vg_quantitative} highlights our approach's ability to effectively expand its understanding of novel concepts with limited samples by effectively utilizing the knowledge graph. Further experiments on the original KG are available in Section~\ref{sec:vgml_quant_more} while a qualitative comparison of the ground-truth graph connections in comparison to the ones \relate adds is available in Section~\ref{sec:vgml_quant_graph_edges}.


\subsubsection{Evaluating Node Addition Procedure}

While \relate allows the addition of novel concepts individually, or as a group, we hypothesize that the node addition strategy has an impact on the overall performance of the model. 
Figure~\ref{fig:onebyonevsallatonce} shows the performance on the VGML dataset when adding the $16$ nodes one-by-one (blue) or all at once (orange), omitting intermediate nodes 6-9 and 11-15 for simplicity. The trends show 
that adding one concept at a time and fine-tuning the classifier as well as the GSNN for each of them before adding the next concept yields a higher performing model. This not only facilitates the extraction and comprehension of new concepts but also prevents the model from getting overwhelmed with multiple concepts simultaneously, thus, minimizing the risk of forgetting previously acquired knowledge.

\subsubsection{Interpretability of Results}
Our approach provides interpretability through the explicit propagation of the initially detected concepts $\sF_\mI$ through the graph $\gG$, providing insights as to why certain final concepts have been classified.
However, while these propagations are not a direct output of the model, they provide an auxiliary insight into the internal workings of the FS-MLC pipeline. Figure~\ref{fig:overview} shows how these propagations can be useful to interpret the concept classifier's result.

\subsection{Limitations and Future Work}
\begin{wrapfigure}{r}{0.45\textwidth}
  \begin{center}
    \includegraphics[width=0.4\textwidth]{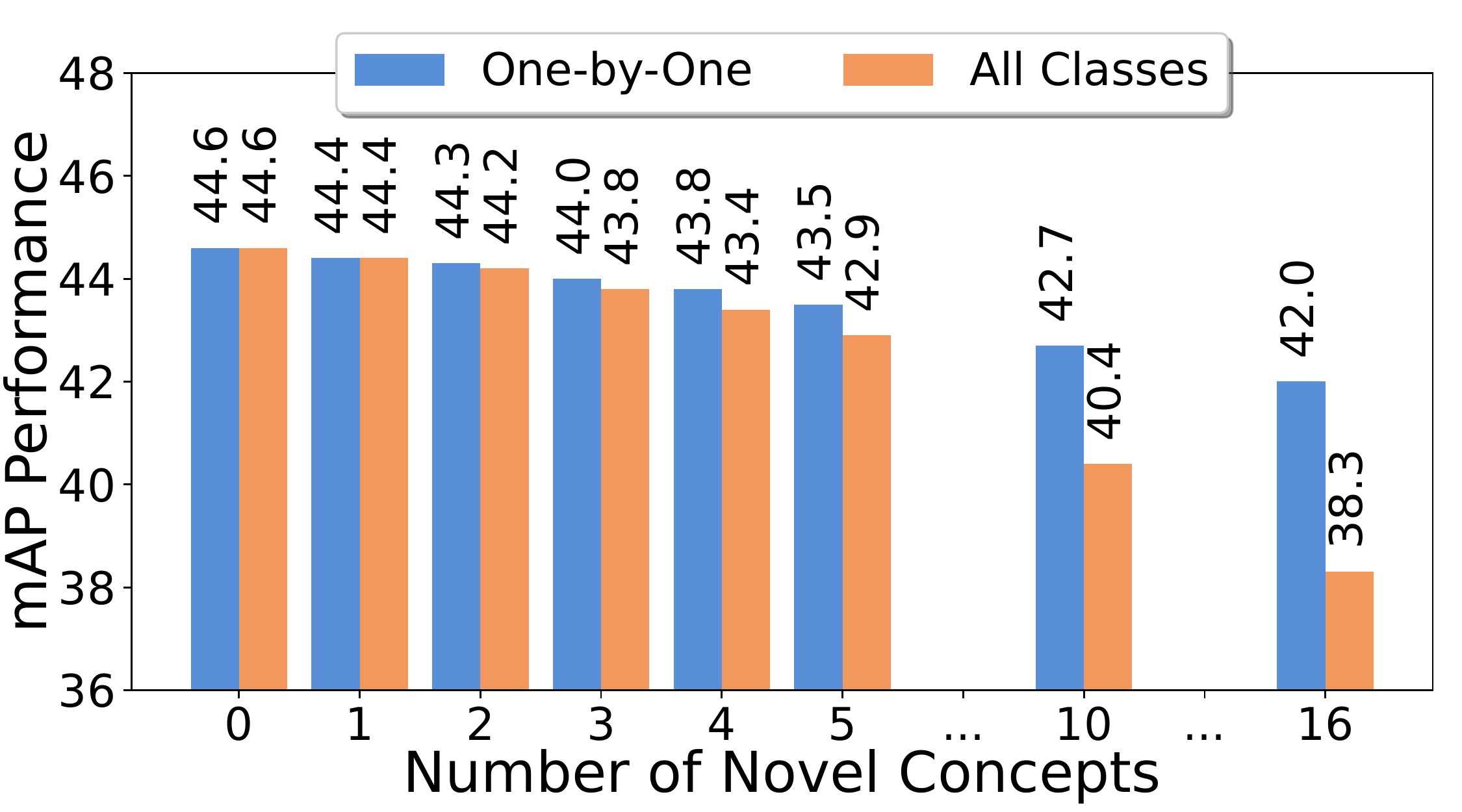}
  \end{center}
  \caption{Strategy for adding the $16$ novel concepts: one-by-one vs. all}
  \label{fig:onebyonevsallatonce}
\end{wrapfigure}
While our method is capable of learning to recognize various objects, abstract concepts, and affordances in a sample-efficient manner, it is dependent on the comprehensiveness of the underlying knowledge graph
Additionally, the reachability of the desired target class from the initially detected concept $\sF_\mI$ depends on the number of propagation steps $T$. 
Further, the accuracy of the model depends on the initial object detections of Faster R-CNN (see Appendix~\ref{sec:obj_detection} for a brief analysis). Another factor adding to this is that \relate requires any potentially related knowledge with respect to a novel concept to be expanded by $\sI_{SME}$ due to the prohibitive computational complexity of checking against every node in $\gG$. In future work, we plan to address these issues by choosing the number of propagation steps dynamically, allowing for further expansions, while also exploring the options of allowing SMEs to review proposed connections that \relate introduces. 

\section{Conclusion}

In this work, we present a novel few-shot multi-label classification approach that uses domain knowledge in the form of a knowledge graph to recognize novel objects, abstract concepts, and affordances. We show that this neuro-symbolic architecture is particularly well suited for few-shot learning as novel concepts can be added to the knowledge graph, and thus alleviates the need to re-train large neural networks, but rather, utilizes a small dataset to perform targeted fine-tuning. 
As part of this methodology, we propose \relate, a novel approach of automatically connecting novel concepts to the existing domain knowledge and show its utility of not only adding novel objects but also adding non-visual concepts. Finally, we show that this approach outperforms current state-of-the-art approaches in few-shot multi-label classification on the COCO dataset. We also show that this approach performs well when on the COCO test classes despite being trained on the VGML dataset, demonstrating the transferability of knowledge through the utilization of a knowledge graph.



\subsubsection*{Acknowledgements}
We gratefully acknowledge the financial support received from multiple funding sources, which made this research possible. Specifically, we would like to acknowledge the support from the Defense Advanced Research Projects Agency (DARPA) under the ASIST grant HR001120C0036, the Air Force Office of Scientific Research (AFOSR) under grants FA9550-18-1-0251 and FA9550-18-1-0097, and the Army Research Laboratory (ARL) under grant W911NF-19-2-0146 and W911NF-2320007.

\clearpage

\bibliography{main}
\bibliographystyle{collas2023_conference}

\newpage

\appendix
\section{Appendix}

\subsection{Cross-Modal Attention Mechanism in \relate} 
\label{sec:cross_attn}


In this section, we elaborate further on the cross-modal attention mechanism to fuse the linguistic concept representation $\vw_c \in \sR^{F_w}$ with the image representation $\mI_P \in \sR^{P \times F^2_PC}$. Fundamentally, this is a standard cross-attention approach in which the word embedding is considered as query and image patch embedding as key and value. However, for completeness, we outline the process as follows.
Particularly, we define
\begin{equation}
    \ve_{c} = f_{MC}\left(\mI_P, \vw_{n}\right) 
\end{equation}

The transformer encoder architecture is built as $L$ sequential layers each composed of a multi-head cross-attention and multi-layer perceptron block where each block is preceded by layer normalization 
and followed by a residual connection.

The initial input to the encoder is a sequence $\tZ^0$ of length $P$ where each element $\vz_p \in \sR^{F_l}$ of size $F_l$ is the computed as follows for each patch in $\mI_{P[i,:]}$:
\begin{equation}
    \vz^0_i = \mI_{P[i,:]}\mE_{[i,:]} + \mP_{[i+1,:]}
\end{equation}
where $\mE \in \sR^{(F_p^2C)\times F_l}$ is a learnable projection matrix and $\mP \in \sR^{(P+1)\times F_l}$ is a learnable positional embedding for each patch in $\mI_P$. Further, we insert a \texttt{CLS} token at the beginning of the list $\tZ_1^0 = \mP_{[0,:]}$.
Provided that the word embedding $\vw_c$ for the concept $c$, for each layer $l \in [1, \dots, L]$, the embedding $\vz_i$ is given by the following equations:
\begin{equation}
    \vz^{l'}_i = f_{CA}\big(\text{layernorm}(\vz_i^{l-1}), \vw_c\big) + \vz^{l-1}_i
\end{equation}
\begin{equation}
    \vz^l_i = \text{MLP}\big(\text{layernorm}(\vz^{l'}_i)\big) + \vz^{l'}_i
\end{equation}

In the above equation, the cross-attention is computed by querying the concept embedding $\vw_c$ against the patchwise encoding of the previous layer, initialized by the patches from image $\mI_P$. 
The cross-attention module, $f_{CA}(\dots)$, is a multi-head approach encompassing $h$ heads. 
Following the standard transformer architecture, we compute $f_{CA}(\dots)$ as follows:
\begin{equation}
    f_{CA}(\vk, \vv, \vq) = \text{softmax}\Big(\frac{\vq\vk^\mathsf{T}}{\sqrt{D_A}}\Big)\vv
\end{equation}


The key $\vk$, query $\vq$, and value $\vv$ for the individual cross-attention heads are given by:
\begin{equation}
    \begin{bmatrix} \vk & \vv \end{bmatrix} = \text{layernorm}(\vz^{l-1}_i) \mW_{kv}, \quad \vq = \vw_c \mW_{q} 
\end{equation}
where $\mW_{kv}$ and $\mW_q$ are trainable weights and $D_A = \frac{F_l}{\beta}$, where $\beta$ is a hyper-parameter.
The final embedding for concept $\ve_{c_n}$ is obtained by extracting the representation corresponding to the $1^{st}$ element in sequence $\tZ$ after all $L$ layers.
\begin{equation}
    \ve_c = f_{MC}\left(\mI_P, \vw_c\right) = \text{layernorm}(\tZ^L_0)
\end{equation}

\subsection{Evaluation of Novel Non-Visual Concept Extraction}
\label{sec:nonvisual_quant}

\begin{table}{}
\centering
\small
\begin{tabular}{C{5.5cm} C{2cm} C{2cm}}  
 \hline
 \textbf{Approach} & \textbf{Affordances} & \textbf{Attributes} \\ [0.5ex]
 \hline
 CLIP (0-shot) & 28.4 & 35.0\\
 Flamingo (0-shot) & 0 & 0\\
 MiniGPT (0-shot) & 18.4 & 24.4\\
 Flamingo (5-shot) & 30.8 & 47.8\\
 Ours (5-shot) & \textbf{61.2} & \textbf{69.8}\\
 \hline
\end{tabular}
\caption{Novel non-visual concept prediction in comparison to free-form text generation models.}

\label{table:nonvisual_quant}
\end{table}

Our model was assessed for its ability to predict non-visual concepts such as affordances and attributes, in comparison to the free-form language generation baselines, namely MiniGPT-4~\citep{zhu2023minigpt4}, Flamingo~\citep{Alayrac2022FlamingoAV}, and CLIP~\citep{Radford2021LearningTV}, explained in Section \ref{sec:nonvisual}. The methodology we utilize to obtain the predictions for each baseline is mentioned below:
\begin{itemize}
    \item CLIP (0-shot) \citep{Radford2021LearningTV}: We evaluated a standard CLIP model by tasking it with a multi-label classification task over our concepts. The language prompt for CLIP is the list of all $316$ nodes plus the novel concept node and we considered the detection to be successful if the targeted concept was part of the $N$ most confident classes, where $N$ is the number of the respective image's ground-truth classes plus one. As CLIP does not provide an easy few-shot learning opportunity, we only evaluated the zero-shot case.
    \item Open-Flaming (0-shot) \citep{Alayrac2022FlamingoAV}: We utilize Open-Flamingo in lieu of the official Flamingo, as official models are not publicly available. However, in the zero-shot case, we provided our test images and prompted Flamingo with the following query: “Does the image show an item that can contain, display, feed, sit, or transport?” for the affordances and "Does the image show an item that is edible, electric, flora, sports, or wearable?" for the attributes. We then evaluated the generated text manually to determine whether or not Flamingo detected the concept correctly. For example, we counted a response like “Yes, the image shows hot dogs with cheese on them, which are edible.” as successful identification of the concept \textit{edible}. 
    \item Open-Flamingo (5-shot) \citep{Alayrac2022FlamingoAV}: In the five-shot use-case, we provided further context to Flamingo by providing all 25 sample images (five for each class) with their respective label to Flamingo and then prompting for a single label for each of the novel test images. 
    \item MiniGPT (0-shot) \citep{zhu2023minigpt4}: Finally, we also employed MiniGPT-4 in order to also utilize a multi-modal GPT baseline. Here, we provided the image as context and asked the same question as in Open-Flamingo while evaluating the generated response manually.
\end{itemize}
The results for the same are presented in Table \ref{table:nonvisual_quant}. Our approach outperforms the best baseline by an average score of $26.2\%$ on the prediction of non-visual concepts. The superior performance of our model can be attributed to its capacity to deduce non-visual concepts by connecting them with visual concepts derived from the visual inputs.


\subsection{Quantitative Evaluation of \relate}
\label{sec:vgml_quant_more}
\setcounter{magicrownumbers}{0}
\setlength{\tabcolsep}{3pt}
\begin{table*}[h]
\centering
\begin{tabular}{c|cc|cc|ccc|ccc} 
 \toprule
 & \multicolumn{2}{c|}{\textbf{Fine-tuned}} & \multicolumn{2}{c|}{\textbf{KG Configuration}} & \multicolumn{3}{c|}{\textbf{All Classes}} & \multicolumn{3}{c}{\textbf{Novel Classes}} \\ 
  & ${\tt GSNN}$ & CLF & ${\tt RelaTe}$ & O-KG & \textbf{T-1} & \textbf{T-5} & \textbf{mAP} & \textbf{T-1} & \textbf{T-5} & \textbf{mAP}\\ 
 \midrule
 \rownumber & - & \checkmark & - &  \checkmark & 90.6 & 70.2 & 38.8 & 91.4 & 72.2 & 67.6 \\
 \rownumber & \checkmark & \checkmark & - & \checkmark & 90.2 & \textbf{72.8} & 41.6 & 91.2 & \textbf{73.7} & 69.0 \\
 \rownumber & - & \checkmark & \checkmark & - & 89.8 & 69.8 & 38.5 & 91.6 & 72.8 & 68.2\\ 
 \rownumber & \checkmark & \checkmark & \checkmark & - & \textbf{90.4} & 72.4 & \textbf{41.7} & \textbf{92.2} & 73.6 & \textbf{69.3} \\ 
 \bottomrule
\end{tabular}
\caption{Experimental results on Visual Genome dataset.}
\label{table:relate_quantitative}
\end{table*}

In addition to the ablations of Table~\ref{table:vg_quantitative}, we provide a quantitative evaluation regarding the ability of \relate to restoring the ground-truth KG of the VGML dataset for the $16$ novel classes. Recall that we intentionally removed the test classes from the KG used in our few-shot experiments. Ideally, \relate~would restore or create an even better KG through the proposed edge-addition framework. 
Table~\ref{table:relate_quantitative} presents a quantitative comparison between our proposed edge addition methodology, \relate, and using the original knowledge graph without removing nodes corresponding to the $16$ novel classes. Rows 1 and 2 demonstrate the use of the original KG (O-KG) while rows 3 and 4 denote the models that use the KG populated by our \relate framework. All four of these models are trained without the use of MDES on a random selection of images from the original dataset.
The results demonstrate that our approach effectively incorporates novel concepts into the KG. In fact, our method outperforms the model that utilized the original KG for some metrics. This is because our approach not only restores the previously removed edges but also introduces additional connections that are observed in the SME-provided images, thus, improving performance. 

\subsection{Qualitative Evaluation of \relate}
\label{sec:vgml_quant_graph_edges}


\begin{figure}
\centering
   \begin{subfigure}{0.49\linewidth} \centering
     \includegraphics[scale=0.45]{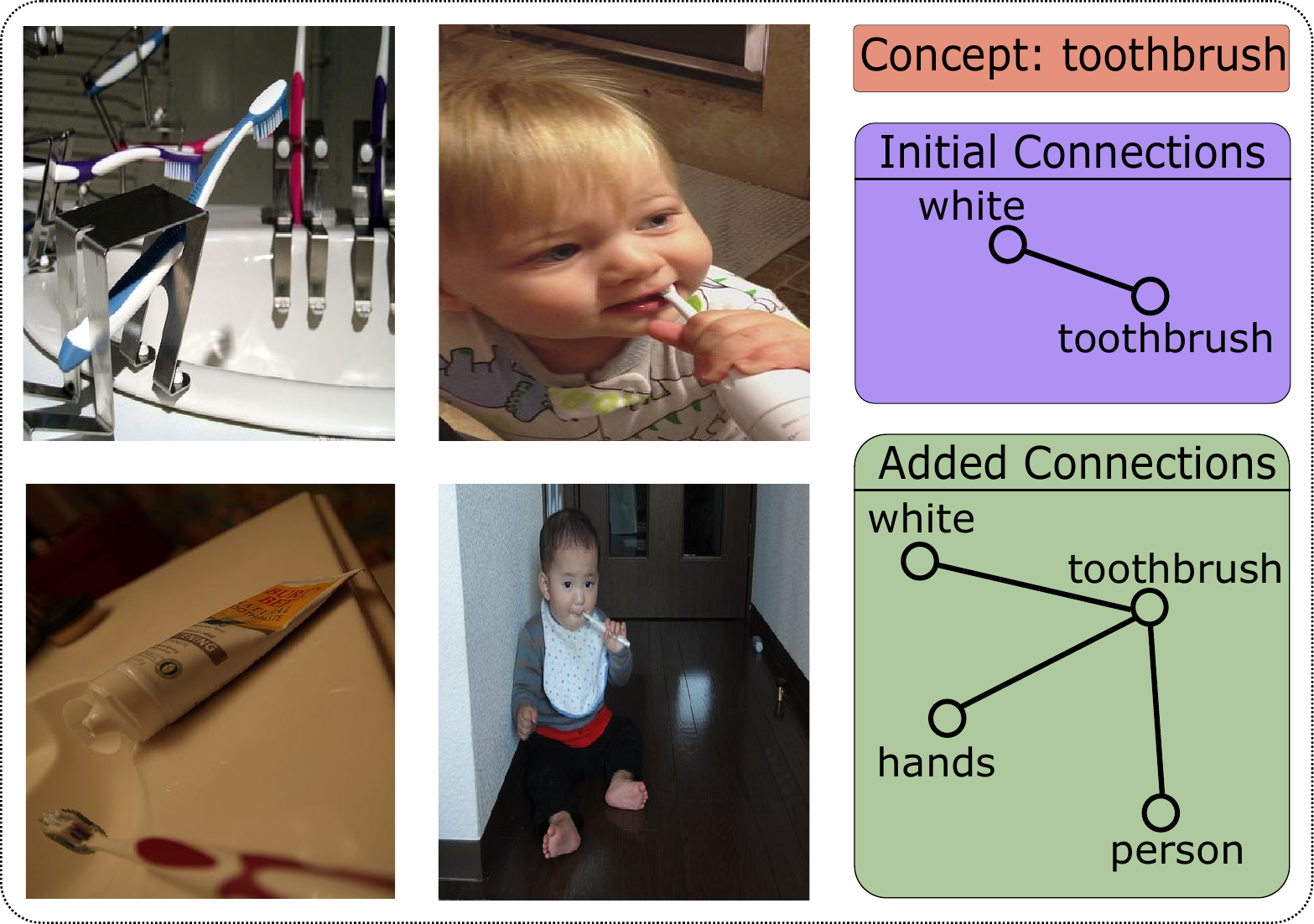}
     \label{fig:relation1}
   \end{subfigure}
    \begin{subfigure}{0.49\linewidth} \centering
     \includegraphics[scale=0.45]{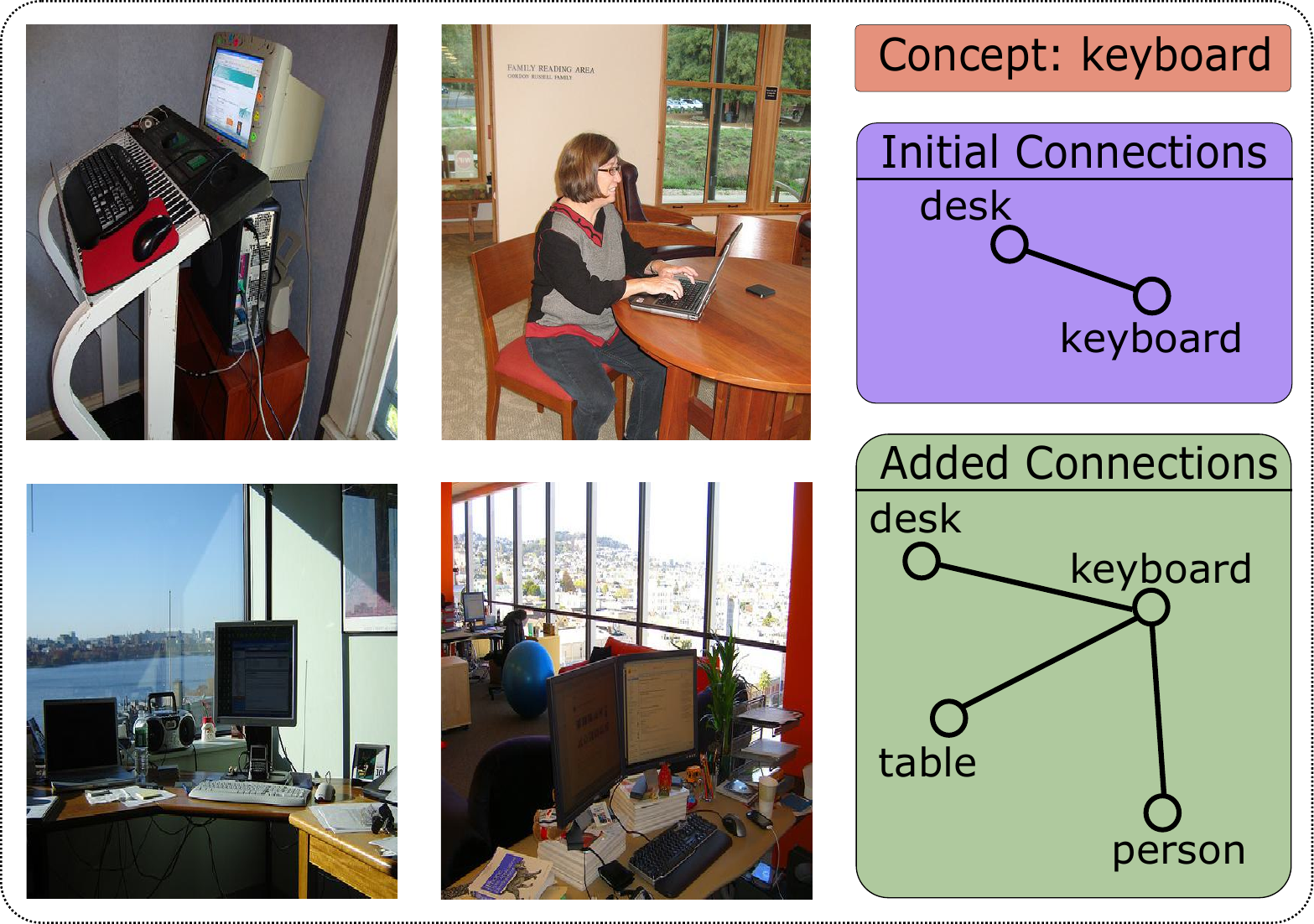}
     \label{fig:relation2}
   \end{subfigure}
    \begin{subfigure}{0.49\linewidth} \centering
     \includegraphics[scale=0.45]{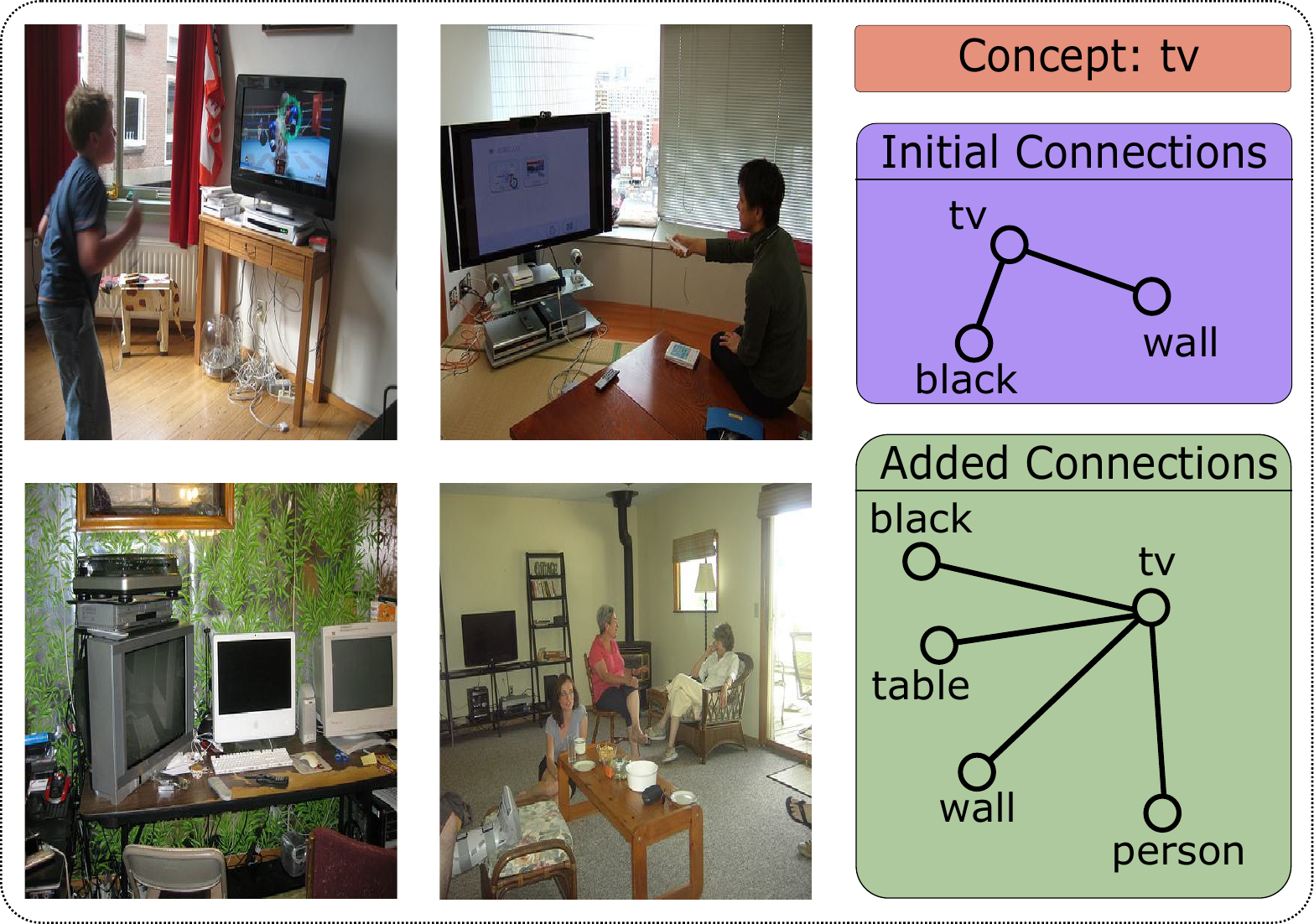}
     \label{fig:relation3}
   \end{subfigure}
   \begin{subfigure}{0.49\linewidth} \centering
     \includegraphics[scale=0.45]{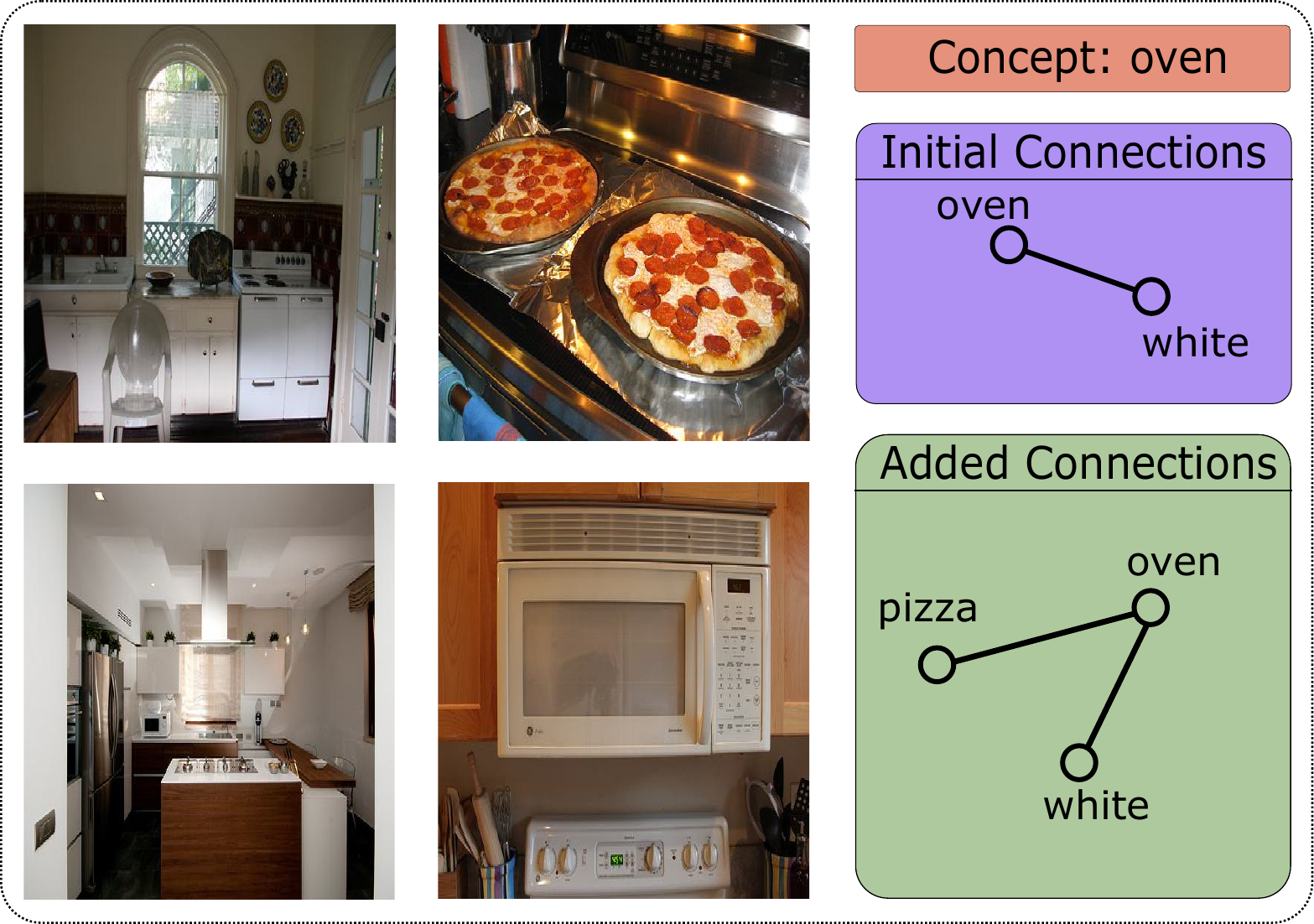}
     \label{fig:relation4}
   \end{subfigure}
\caption{Qualitative Evaluation of Edges added by the \relate~approach. In each example, we include the concept being added, the edges that were present in the knowledge graph originally, and the nodes that were suggested by \relate~for a set of images.} \label{fig:relate_examples}
\end{figure}
In this section, we provide examples of connections recommended by our \relate framework. For each novel concept, we pick $4$ images and pass them through our edge addition framework to demonstrate the edges that it populates into the graph $\gG$. 
In Figure \ref{fig:relate_examples}, each example starts with the concept that was removed from the knowledge graph (red) and its initial connections (purple). The suggested connections by our \relate framework are shown in the green box, which was generated when the system was given a set of $4$ images. These results demonstrate the effectiveness of our relation prediction approach in adding back relevant connections that are prominent in the provided images. Moreover, our model suggests some additional relations that may not have been present in the original graph but are relevant and can provide significant information about the scene content.

\subsection{Countering the Classifier Bottleneck}

\begin{wrapfigure}{r}{0.5\textwidth}
  \begin{center}
    \includegraphics[width=0.45\textwidth]{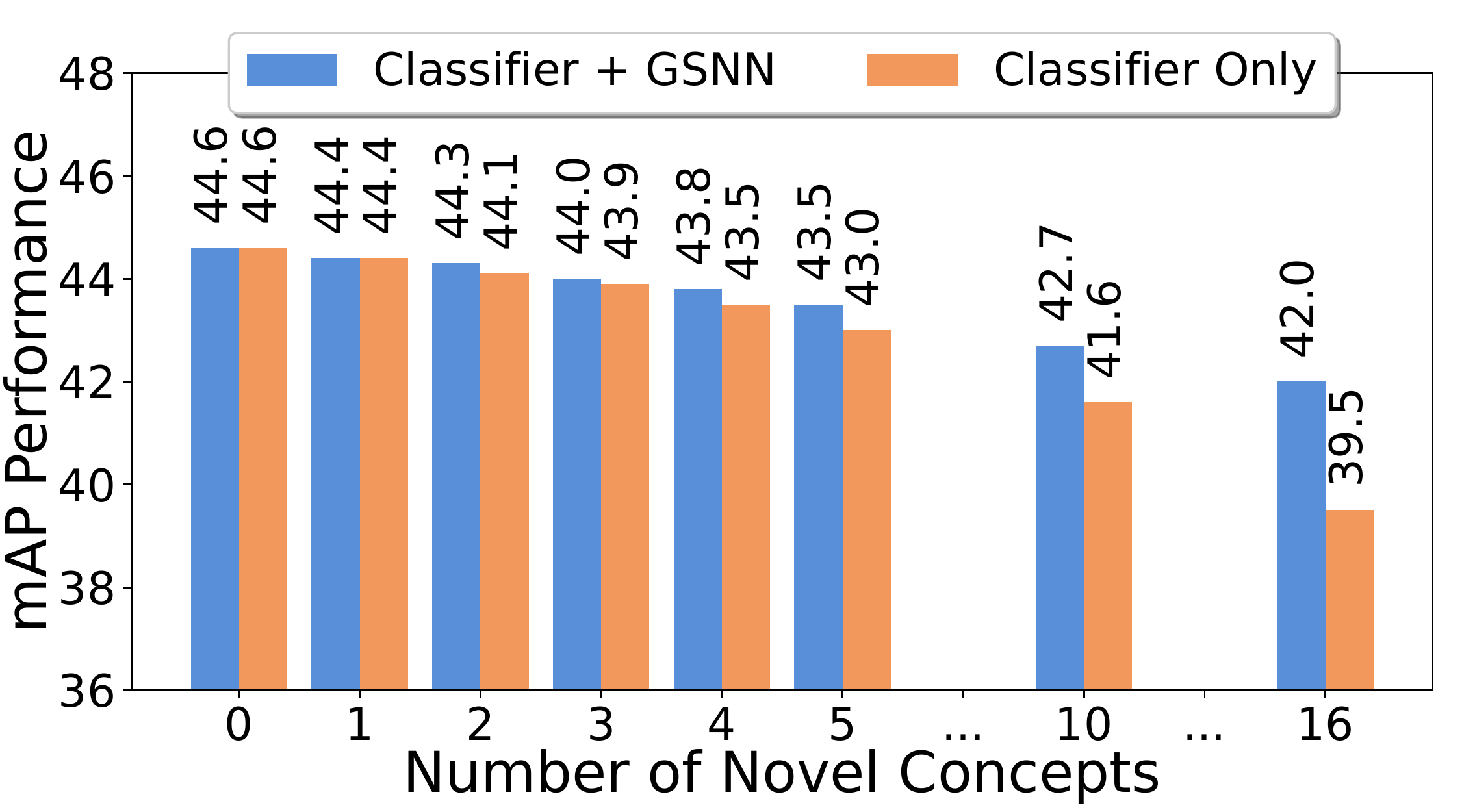}
  \end{center}
  \caption{Fine-tuning GSNN + Classifier vs Classifier Only}
  \label{fig:clsvsgsnn}
\end{wrapfigure}

We demonstrate the crucial role of fine-tuning the propagation network and the node bias when adding novel concepts to the graph. In Figure~\ref{fig:clsvsgsnn}, we plot the mAP performance on the entire VGML dataset as a function of the number of novel classes added to the system for the model where we fine-tune either only the classification or both classification and the propagation module including the node biases. During training with five images per concept, we utilize the one-by-one node addition strategy which showed improved performance (see Figure~\ref{fig:onebyonevsallatonce}). Initially, for just a few nodes, not training the GSNN does not have a huge influence; however, the plot shows that the model in which we only fine-tune the classifier experiences a substantial performance drop which is proportional to the number of concepts added compared to the model in which both the modules are fine-tuned. 



\subsection{Ablation on Number of Propagation Steps} 
\label{sec:iterations}

\begin{table}{}
\centering
\small
\begin{tabular}{C{2.5cm} C{2cm} C{2cm}}  
 \hline
 \textbf{Steps of Expansion, $T$} & \textbf{Expansion \%} & \textbf{mAP} \\ [0.5ex]
 \hline
 2 & 39.5 & 39.1\\
 3 & 93.4 & \textbf{42.0}\\
 4 & 100.0 & 41.3\\
 \hline
\end{tabular}
\caption{Percentage of samples that required $T$ steps of expansion and the corresponding mAP performance of our model with that $T$.}
\label{table:steps_expansion}
\end{table}

We experiment with different values for $T$ that define the number of iterations between the propagation and importance network during inference of the GSNN module as described in Section~\ref{sec:GSNN}. We aim to select the minimum possible value of $T$ that ensures the complete expansion of most of the samples in our test dataset within the first $T$ steps. In Table~\ref{table:steps_expansion}, we report the performance of our model on all the classes of the VGML test dataset along with the percentage of samples that were expanded to full capacity by varying the number of expansion steps.
The results we obtained in Table~\ref{table:steps_expansion} highlight that $3$ is the optimal value of $T$ since we start to obtain diminishing returns following steps greater than $3$. Expanding less than two steps doesn't allow the model to experience many relevant connections while expanding beyond the third level makes it challenging for the model to identify concepts that are related to the original concepts $\sF_\mI$.

\subsection{Significance of Image Conditioning on Node Embeddings}
\label{sec:image_conditioning}
We explicitly enforce conditioning of the image content on the embeddings generated for each of the nodes during the graph propagation. These embeddings are utilized by both the importance and the context network and form the backbone of the entire graph expansion procedure. 
Unlike~\citet{Marino2016TheMY}, which does not enforce this constraint, our model prioritizes expanding nodes that are relevant to the image content rather than simply expanding nodes that are only dependent on the initial class detection that would result in the same propagated nodes even for dissimilar images.
\begin{figure}[h]
\centering
   \begin{subfigure}{0.3\linewidth} \centering
     \includegraphics[width=4cm,height=4cm]{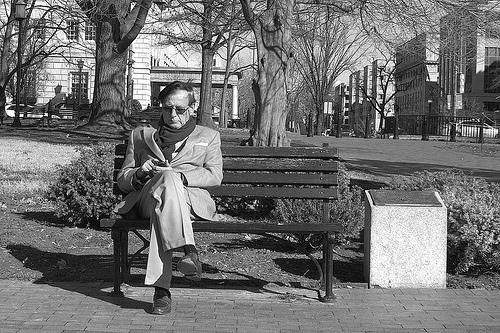}
     \label{fig:eg1}
  \end{subfigure}
     \begin{subfigure}{0.3\linewidth} \centering
     \includegraphics[width=4cm,height=4cm]{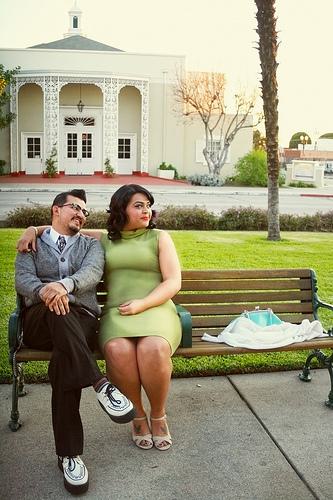}
     \label{fig:eg2}
  \end{subfigure}
     \begin{subfigure}{0.3\linewidth} \centering
     \includegraphics[width=4cm,height=4cm]{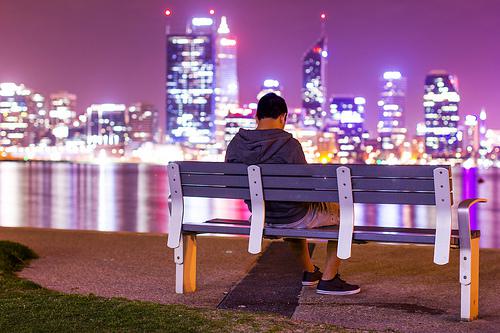}
     \label{fig:eg3}
  \end{subfigure}
\caption{Samples from the dataset where the initial propagation begins with the concepts of \textit{person} and \textit{bench}.} \label{fig:conditioning_imp}
\end{figure}
We demonstrate the significance of our proposed conditioning by selecting three vastly distinct images from the dataset in Figure~\ref{fig:conditioning_imp}. The original GSNN fails to distinguish between these images in terms of expanded nodes in the KG, whereas our approach expands a unique set of nodes for each image. The following are the final classifications with and without image conditioning on the propagation network:
\begin{itemize}
    \item Image 1: 
        \begin{itemize}
            \item w/o Conditioning: \concept{person}, \concept{bench}, \concept{shirt}, \concept{black}, \concept{white}, \concept{gray}
            \item w Conditioning: \concept{person}, \concept{bench}, \concept{shirt}, \concept{wooden}, \concept{brown}, \concept{black}, \concept{sunglasses} 
        \end{itemize}
    \item Image 2: 
        \begin{itemize}
            \item w/o Conditioning: \concept{person}, \concept{bench}, \concept{green}, \concept{sitting}, \concept{shirt}, \concept{white}
            \item w Conditioning: \concept{person}, \concept{bench}, \concept{jacket}, \concept{green}, \concept{visible}, \concept{sitting}
        \end{itemize}
    \item Image 3: 
        \begin{itemize}
            \item w/o Conditioning: \concept{person}, \concept{bench}, \concept{shirt}, \concept{sitting}, \concept{pink}, \concept{wooden}, \concept{black}
            \item w Conditioning: \concept{person}, \concept{bench}, \concept{shirt}, \concept{black}, \concept{jacket}, \concept{head}, \concept{wooden} 
        \end{itemize}
\end{itemize}

While the model without image conditioning expands a generic list of nodes, our approach identifies image-specific concepts such as \textit{sunglasses} for the first image and \textit{jacket} for the second image, demonstrating the improvements imposed by this additional conditioning.

\subsection{Ablation for Node types and Edge types}
\label{sec:node_edge_ablation}

\setcounter{magicrownumbers}{0}
\begin{minipage}[t]{0.49\textwidth}
    \centering
    \begin{tabular}{cccc}  
        \toprule
         & \textbf{Edge types} & \textbf{Node types} & \textbf{mAP} \\ 
        \midrule
         \rownumber & \checkmark & - & 42.3 \\
         \rownumber & - & - & 42.1 \\
         \rownumber & - & \checkmark & \textbf{44.6} \\
         \bottomrule
        \end{tabular}
      \captionof{table}{Experimental results with ablations of edge and node types.}
      \label{table:edge_node_ablation_table}
    \end{minipage}
    \hspace{0.015\textwidth}
\begin{minipage}{\textwidth}
  \begin{minipage}[b]{0.49\textwidth}
    \centering
    \includegraphics[width=1\linewidth]{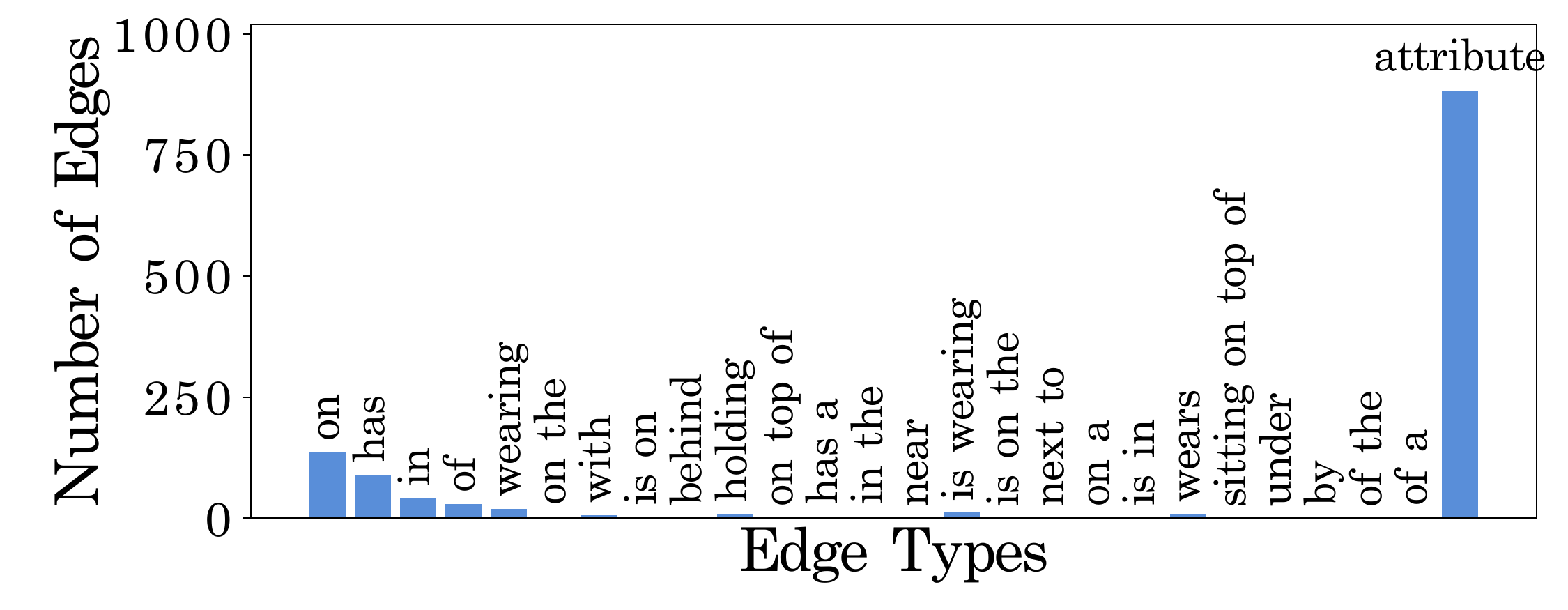}
    \captionof{figure}{Edge type distribution of the KG used by \cite{Marino2016TheMY}.}
    \label{fig:edgetype_distribution}
  \end{minipage}
  \hfill
  \end{minipage}

In Section~\ref{sec:data}, we introduced the changes to the KG as described in~\citet{Marino2016TheMY}. Here, we ablate these choices in greater detail. Table~\ref{table:edge_node_ablation_table} shows the performance of the algorithm with the original $26$ edge types in line 1, no edge types (i.e. just a single unlabeled edge) in line 2, and our modified KG without edge types, but an additional one-hot indicating the node-type in line 3. The results show that edge types hinder the performance of the inference pipeline and indicating the node type improves performance. We hypothesize that this is due to the strong imbalance of the encoded edge types, as shown in Figure~\ref{fig:edgetype_distribution}, where the \textit{has attribute}, comprising almost two-thirds of all edge types.

\subsection{Maximally Diverse Expansion Sampling}
\label{sec:mdes}

To select a small subset of the original dataset that allows us to maximize the diversity of expanded nodes in our KG, we adopt a binning-based approach. We begin with a single bin spanning all nodes and traverse the dataset to identify the image that can expand a node in the largest bin. Upon finding such an image, we use each expanded node in that image as the dividing line between the new bins. If an image does not expand a node that would divide the largest bin, the image is not added to our curated dataset $\train_C$. We only process the dataset once until either we have a set of images that expand all the possible nodes or all images have been either added to $\train_C$ or have been discarded. Under the assumption that rare classes are randomly distributed in the dataset, we ensure that at least some images containing that class are added to $\train_C$. As a result, we create a dataset $\train_C$ containing approximately $2\%$ of the original VGML dataset.

\subsection{Analysis of Dependence on Object Detectors}
\label{sec:obj_detection}

\begin{figure}
\centering
   \begin{subfigure}{0.45\linewidth} \centering
     \includegraphics[width=1\linewidth]{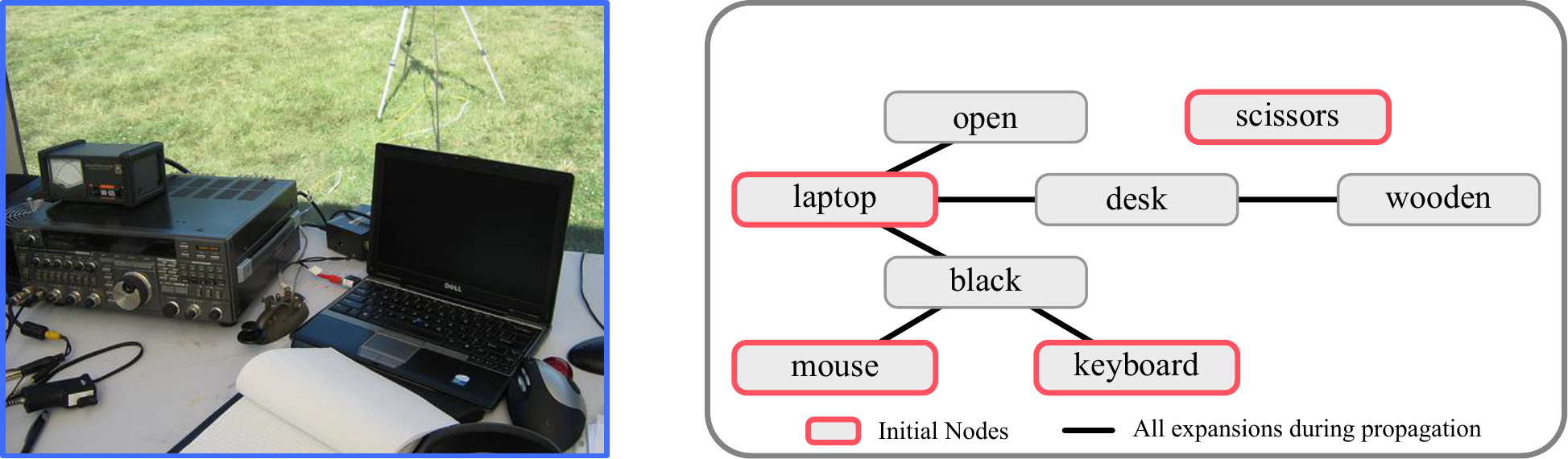}
     \label{fig:fc_3}
     \caption{Mitigated failure case: While Faster R-CNN detected a \textit{scissors}, our propagation and importance network did not incorporate this node any further.}
  \end{subfigure}
  \hspace{0.06\linewidth}
     \begin{subfigure}{0.45\linewidth} \centering
     \includegraphics[width=1\linewidth]{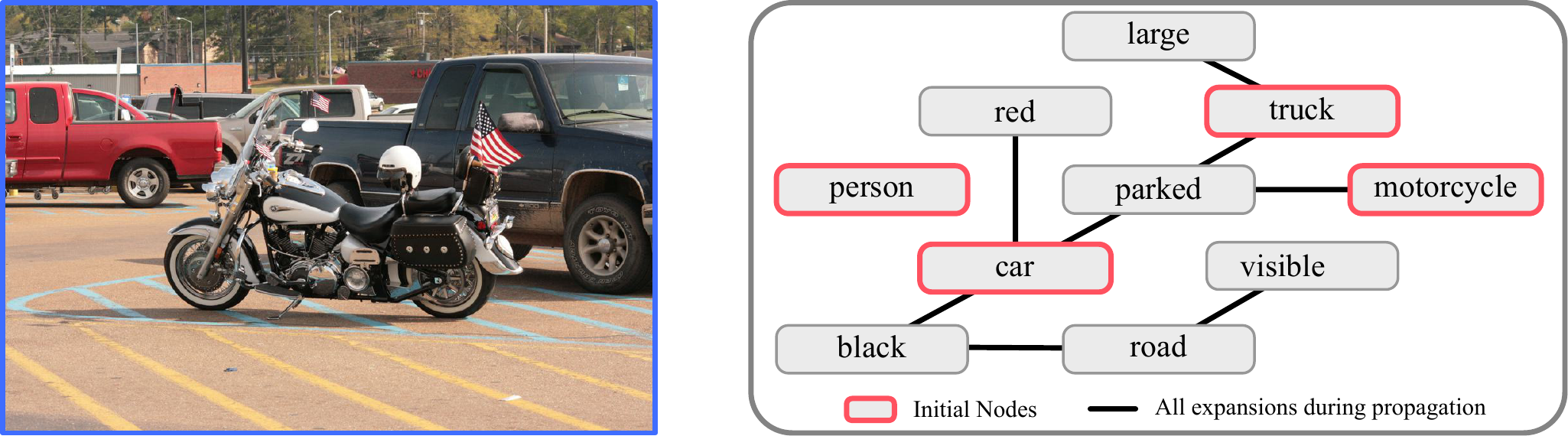}
     \label{fig:fc_4}
     \caption{Mitigated failure case: While Faster R-CNN detected a \textit{person}, our propagation and importance network did not incorporate this node any further.}
  \end{subfigure}
\caption{Robustness to wrong graph initialization by Faster R-CNN detections.} \label{fig:failure_cases_2}
\end{figure}

\begin{figure}
\centering
   \begin{subfigure}{0.45\linewidth} \centering
     \includegraphics[width=1\linewidth]{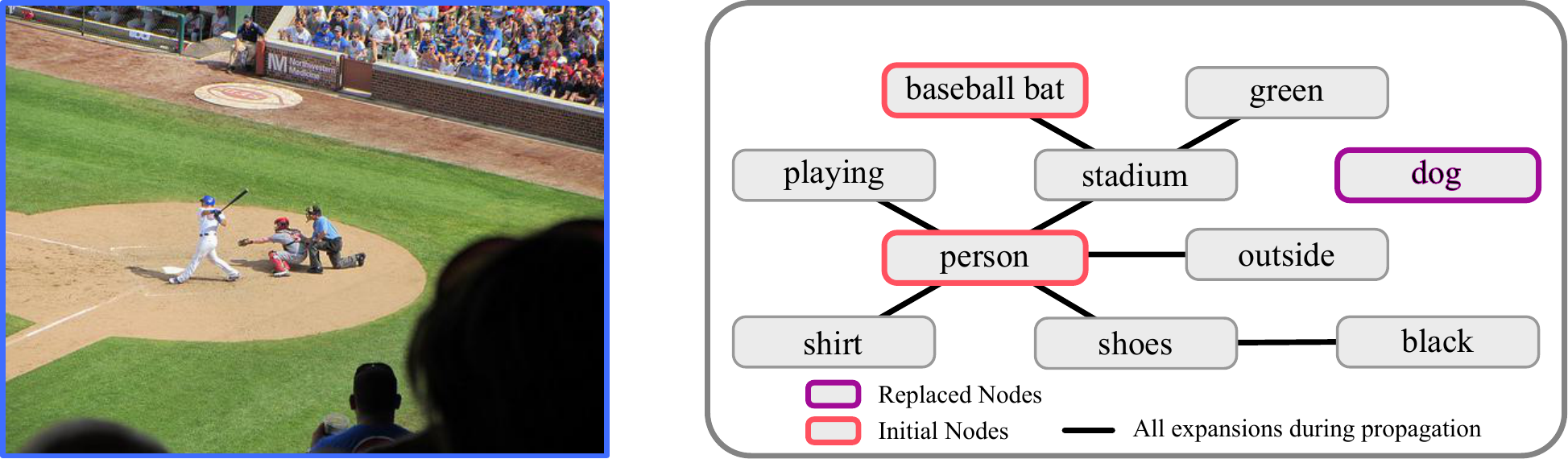}
     \label{fig:rep_1}
     \caption{Node Replacement: Here, we removed the \textit{sports ball} and replaced it with a \textit{dog}, demonstrating how our approach does not incorporate the wrong node.}
  \end{subfigure}
  \hspace{0.06\linewidth}
     \begin{subfigure}{0.45\linewidth} \centering
     \includegraphics[width=1\linewidth]{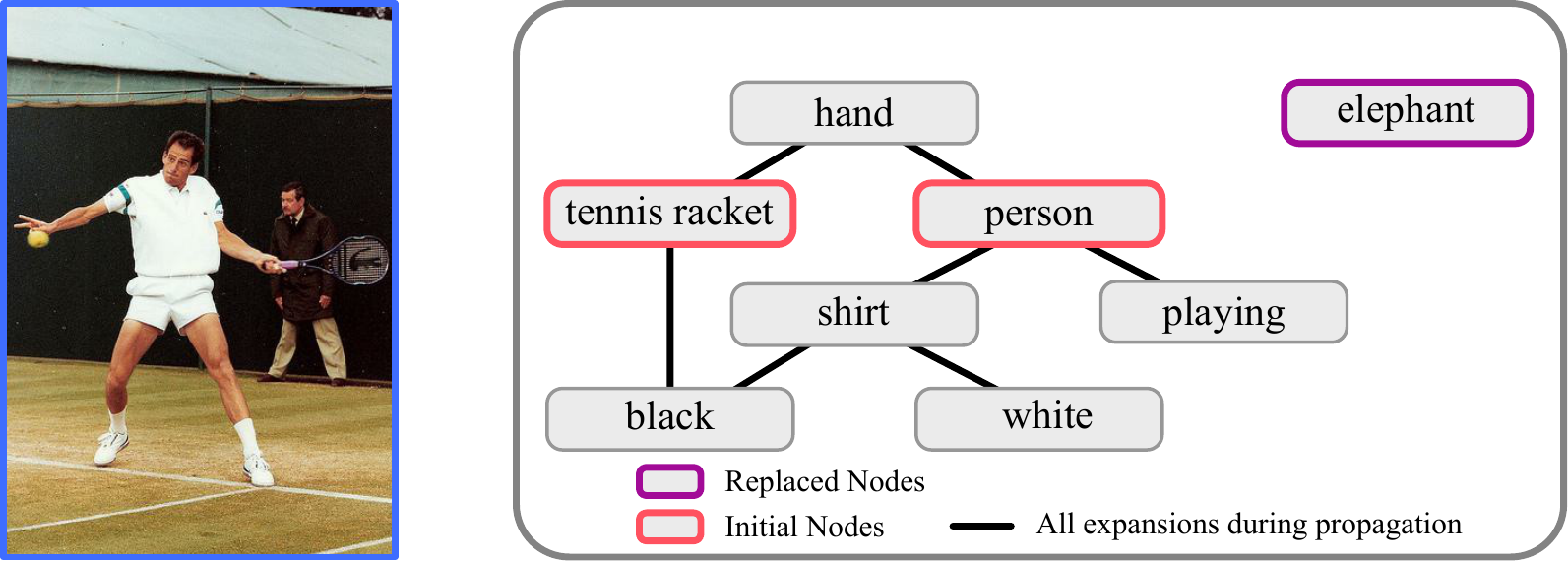}
     \label{fig:rep_2}
     \caption{Node Replacement: Here, we replaced the \textit{sports ball} with an \textit{elephant}. An interesting result here is that not only was the wrong node not expanded, but it was also removed from the final classification.}
  \end{subfigure}
\caption{Robustness to wrong graph initializations that are manually enforced.} \label{fig:replacements}
\end{figure}

To test the resilience of our model against inaccuracies in the object detection module, we conducted an evaluation by replacing the detected objects with random concepts (that were not originally present in the respective example), and observing whether our model expands upon them. 
We conducted a small-scale experiment on $30$ test images, where we introduced an additional random node that is unrelated to the actual image. We observed that the propagation and importance networks ignore these wrong nodes in $63\%$ of the cases by not expanding them any further. Further, in $16.7\%$ of the cases, 
the final classifier removes these nodes altogether. In the current work, the importance network can not remove previously added nodes; however, this capability could be explored in future work. 
Figure~\ref{fig:failure_cases_2} demonstrates two instances where Faster R-CNN mistakenly detects a non-existent object class in the image.
Furthermore, we provide a few instances in Figure \ref{fig:replacements} where we substitute one of the original object detections in the image with an entirely unrelated object, and our model refrains from further propagating the modified node, demonstrating its resistance to such potential issues.

Additionally, we also analyzed a potential failure case in which wrong edges exist in the graph. The only potential source for such edges is if \relate predicts wrong edges during novel concept addition. When testing the performance of \relate by removing a known node for which the desired edges are known, we observe that $84\%$ of these edges are restored when re-adding the target node using our approach. However, while the remaining $16\%$ of edges are not necessarily wrong, we analyzed the impact of potentially wrong edges by manually introducing them between the initially detected nodes and an arbitrary, unrelated node. This was evaluated on $30$ images, as for the prior experiments. We observe that in $76.6\%$ of the cases, the propagation and importance network ignore this wrong connection.

This highlights the robustness of our model to erroneous initialization. Moreover, we empirically observed that Faster R-CNN rarely introduces wrong nodes, thus further mitigating this potential error source.


\subsection{Failure Analysis}
\label{sec:failure}


\begin{figure}
\centering
   \begin{subfigure}{0.45\linewidth} \centering
     \includegraphics[width=1\linewidth]{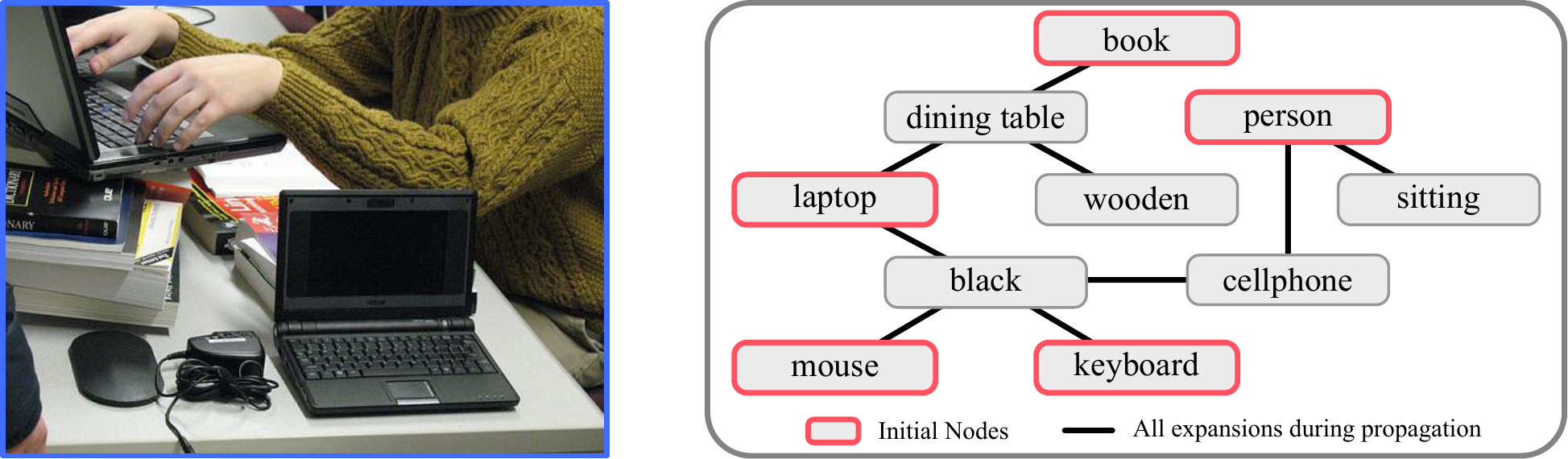}
     \caption{Failure case: The model identified an erroneously identified and further integrated a \textit{cellphone}.}
     \label{fig:fc_1}
  \end{subfigure}
  \hspace{0.06\linewidth}
     \begin{subfigure}{0.45\linewidth} \centering
     \includegraphics[width=1\linewidth]{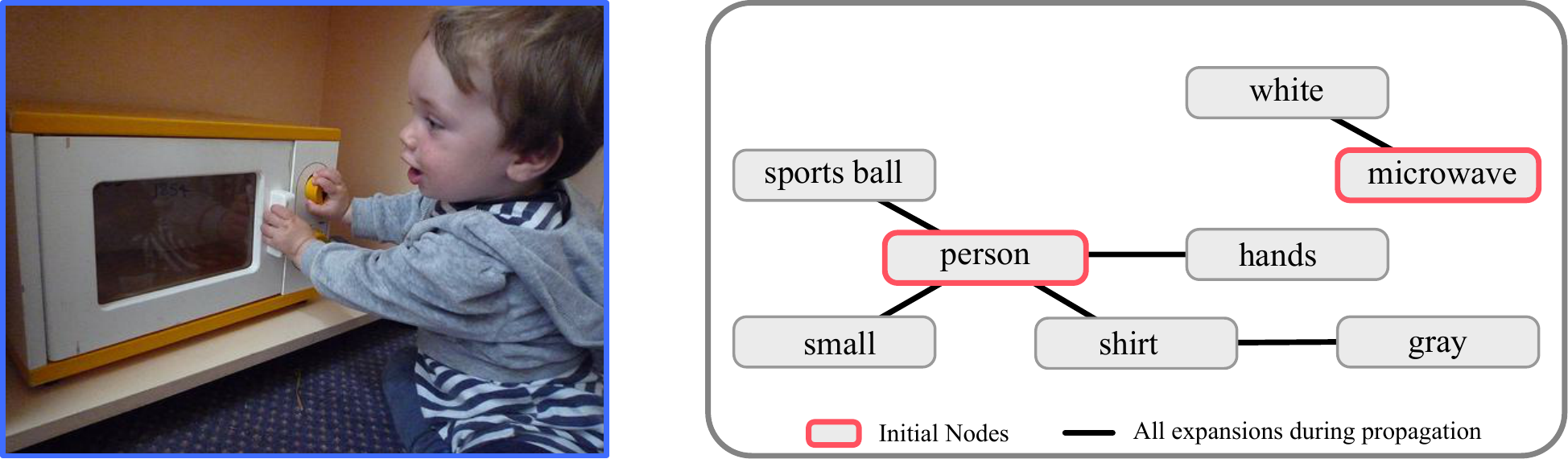}
     \caption{Failure case: The model identified an erroneously identified and further integrated a \textit{sports ball}.}
     \label{fig:fc_2}
  \end{subfigure}
\caption{Failure cases of our model in which wrong nodes are integrated into the graph.} \label{fig:failure_cases}
\end{figure}

As part of our failure analysis, we highlight some examples where our model hallucinates non-existent concepts in the image. Although such misclassifications are not common, they offer valuable insights into how our approach functions and where it may be prone to errors. \\
In the given scenario depicted in Figure~\ref{fig:fc_1}, the model has incorrectly identified the concept of a \textit{cellphone}. This error can be attributed to the model's tendency to associate objects with certain visual characteristics, which can result in confusion between objects that share common properties. For instance, in this case, both the \textit{laptop} and \textit{cellphone} have a screen, and therefore the affordance of being able to display something, leading to the misidentification of the object as a \textit{cellphone}.
The second example in Figure \ref{fig:fc_2} demonstrates another instance where our model has made an incorrect prediction by identifying the object in the image as a \textit{sports ball}. This error can be attributed to the model's tendency to rely on the way people interact with objects in the scene when identifying them. In this case, the child's hand gripping the knob of the \textit{microwave} may resemble the way one would grip a ball, leading the model to mistakenly classify it as a \textit{sports ball}.

Finally, we evaluate potential failure cases in which \relate may be tasked to add edges between contextually unrelated nodes. It is a key feature of \relate to automatically determine the nodes that are relevant for a novel concept while not adding edges to nodes that are contextually different. To evaluate this, we attempt to add edges between nodes from the \textit{bedroom} context and nodes from the \textit{stadium} context. In this case, we observe that \relate only adds an edge in $20\%$ of the queried connections. However, it is important to note that some connections are in fact reasonable, as connections between the \textit{person} node in the stadium context have a valid connection to \textit{bed} in the bedroom context.

\subsection{Runtime Complexity}
We trained our model on a single RTX 6000 GPU for $\approx100$ hours of total training time. When adding a novel concept, the two-staged tuning of our model takes approximately $45$ minutes. Finally, during inference, it takes approximately $30$ seconds per image to obtain predictions using our approach. 

\end{document}